\documentclass[10pt,twocolumn,letterpaper]{article}

\usepackage[pagenumbers]{cvpr} %

\usepackage{graphicx}
\usepackage{amsmath}
\usepackage{amssymb}
\usepackage{booktabs}
\usepackage{makecell}
\usepackage{numprint}
\usepackage{algorithm2e}
\usepackage{float}
\newtheorem{definition}{Def.}
\usepackage{amstext} %
\usepackage{array}   %
\newcolumntype{L}{>{$}l<{$}} %
\newcolumntype{C}{>{$}c<{$}} %

\usepackage[pagebackref,breaklinks,colorlinks]{hyperref}

\usepackage[capitalize]{cleveref}
\Crefname{section}{Sec.}{Secs.}
\Crefname{section}{Section}{Sections}
\Crefname{table}{Table}{Tables}

\newcommand\boldred[1]{\textcolor{red}{\mathbf{#1}}}
\newcommand\red[1]{\textcolor{red}{#1}}

\def\cvprPaperID{104} %
\def\confName{CVPR}
\def\confYear{2022}

\DeclareMathOperator*{\argmax}{arg\,max}

\begin{document}

\title{Rethinking Efficient Lane Detection via Curve Modeling}

\author{Zhengyang Feng$^{1*}$,~~~Shaohua Guo$^{1}$\thanks{Equal Contribution.},~~~Xin Tan$^{2,1}$,~~~Ke Xu$^{3}$,~~~Min Wang$^{4}$,~~~Lizhuang Ma$^{1,2,5}$\thanks{Lizhuang Ma is a member of Qing Yuan Research Institute, Shanghai Jiao Tong University.}\\
$^1$Shanghai Jiao Tong University~~~$^2$East China Normal University~~~$^3$City University of Hong Kong\\$^4$SenseTime Research~~~$^5$MoE Key Lab of Artificial Intelligence, Shanghai Jiao Tong University\\
\tt\small zyfeng97@sjtu.edu.cn;guoshaohua@sjtu.edu.cn;tanxin2017@sjtu.edu.cn;kkangwing@gmail.com;\\
\tt\small wangmin@sensetime.com;ma-lz@cs.sjtu.edu.cn
}

\maketitle

\begin{abstract}
   This paper presents a novel parametric curve-based method for lane detection in RGB images.
   Unlike state-of-the-art segmentation-based and point detection-based methods that typically require heuristics to either decode predictions or formulate a large sum of anchors, the curve-based methods can learn holistic lane representations naturally.
   To handle the optimization difficulties of existing polynomial curve methods, we propose to exploit the parametric Bézier curve due to its ease of computation, stability, and high freedom degrees of transformations.
   In addition, we propose the deformable convolution-based feature flip fusion, for exploiting the symmetry properties of lanes in driving scenes.
   The proposed method achieves a new state-of-the-art performance on the popular LLAMAS benchmark. It also achieves favorable accuracy on the TuSimple and CULane datasets, while retaining both low latency (\textgreater 150 FPS) and small model size (\textless 10M). 
   Our method can serve as a new baseline, to shed the light on the parametric curves modeling for lane detection. Codes of our model and PytorchAutoDrive: a unified framework for self-driving perception, are available at: \url{https://github.com/voldemortX/pytorch-auto-drive} .
\end{abstract}

\section{Introduction}
\label{sec:intro}

\begin{figure}[t]
    \centering
    \includegraphics[scale=1]{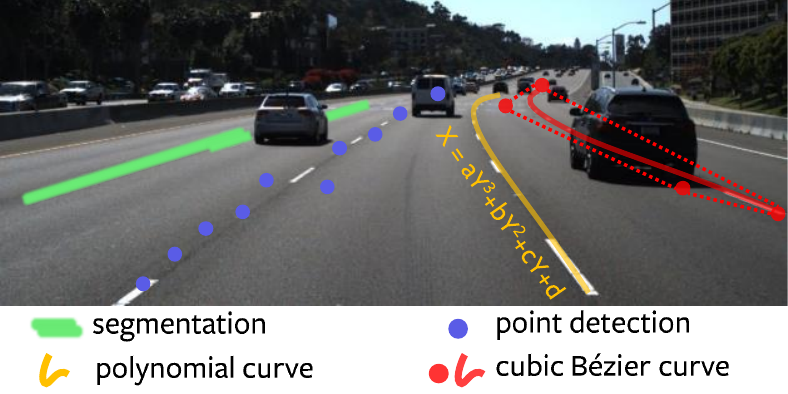}
    \vspace{-3mm}
    \caption{Lane detection strategies. Segmentation-based and point detection-based representations are local and indirect. The abstract coefficients (a, b, c, d) used in polynomial curve are hard to optimize.
    The cubic Bézier curve is defined by 4 actually existing control points, which roughly fit line shape and wrap the lane line in its convex hull (dashed red lines). Best viewed in color.}
    \label{fig:modeling}
    \vspace{-3mm}
\end{figure}

Lane detection is a fundamental task in autonomous driving systems, which supports the decision-making of lane-keeping, centering and changing, \etc.
Previous lane detection methods~\cite{badue2021self,hillel2014recent} typically rely on expensive sensors such as LIDAR.
Advanced by the rapid development of deep learning techniques, many works~\cite{pan2018spatial,neven2018towards,zheng2021resa,tabelini2021keep,liu2021end} are proposed to detect lane lines from RGB inputs captured by commercial front-mounted cameras.

Deep lane detection methods can be classified into three categories, \ie, segmentation-based, point detection-based, and curve-based methods (\Cref{fig:modeling}).
Among them, by relying on classic segmentation \cite{deeplabv1} and object detection \cite{ren2015faster} networks, the segmentation-based and point detection-based methods typically achieve state-of-the-art lane detection performance.
The segmentation-based methods~\cite{pan2018spatial,zheng2021resa,neven2018towards} exploit the foreground texture cues to segment the lane pixels and decode these pixels into line instances via heuristics.
The point detection-based methods~\cite{li2019line,tabelini2021keep,xu2020curvelane} typically adopt the R-CNN framework~\cite{girshick2015fast,ren2015faster}, and detect lane lines by detecting a dense series of points (\eg, every 10 pixels in the vertical axis).
Both kinds of approaches represent lane lines via indirect proxies (\ie, segmentation maps and points).
To handle the learning of holistic lane lines, under cases of occlusions or adverse weather/illumination conditions, they have to rely on low-efficiency designs, such as recurrent feature aggregation (too heavy for this real-time task)~\cite{pan2018spatial,zheng2021resa}, or a large number of heuristic anchors ($>1000$, which may be biased to dataset statistics)~\cite{tabelini2021keep}. %

On the other hand, there are only a few methods~\cite{tabelini2021polylanenet,liu2021end} proposed to model the lane lines as holistic curves (typically the polynomial curves, \eg, $x = ay^3 + by^2 + cy + d$).
While we expect the holistic curve to be a concise and elegant way to model the geometric properties of lane line, the abstract polynomial coefficients are difficult to learn. Previous studies show that their performance lag behind the well-designed segmentation-based and point detection-based methods by a large margin (up to $8\%$ gap to state-of-the-art methods on the CULane \cite{pan2018spatial} dataset).
\textit{In this paper, we aim to answer the question of whether it is possible to build a state-of-the-art curve-based lane detector.
}

We observe that the classic cubic Bézier curves, with sufficient freedom degrees of parameterizing the deformations of lane lines in driving scenes, remain low computation complexity and high stability.
This inspires us to propose to model the thin and long geometric shape properties of lane lines via Bézier curves. 
The ease of optimization from on-image Bézier control points enables the network to be end-to-end learnable with the bipartite matching loss~\cite{wang2021end}, using a sparse set of lane proposals from simple column-wise Pooling (\eg, 50 proposals on the CULane dataset~\cite{pan2018spatial}), without any post-processing steps such as the Non-Maximum Suppression (NMS), or hand-crafted heuristics such as anchors, hence leads to high speed and small model size.
In addition, we observe that lane lines appear symmetrically from a front-mounted camera (\eg, between ego lane lines, or immediate left and right lanes). To model this global structure of driving scenes,
we further propose the feature flip fusion, to aggregate the feature map with its horizontally flipped version, to strengthen such co-existences. 
We base our design of feature flip fusion on the deformable convolution~\cite{zhu2019deformable},
for aligning the imperfect symmetries caused by, \eg, rotated camera, changing lane, non-paired lines.
We conduct extensive experiments to analyze the properties of our method and show that it performs favorably against state-of-the-art lane detectors on three popular benchmark datasets. Our main contributions are summarized as follows:

\begin{itemize}
    \item We propose a novel Bézier curve-based deep lane detector, which can model the geometric shapes of lane lines effectively, and be naturally robust to adverse driving conditions.
    \item We propose a novel deformable convolution-based feature flip fusion module, to exploit the symmetry property of lanes observed from front-view cameras.
    \item We show that our method is fast, light-weight, and accurate through extensive experiments on three popular lane detection datasets.
    Specifically, our method outperforms all existing methods on the LLAMAS benchmark~\cite{llamas2019}, with the light-weight ResNet-34 backbone.
\end{itemize}

\section{Related Work}
\label{sec:relat}

\noindent \textbf{Segmentation-based Lane Detection.} These methods represent lanes as per-pixel segmentation. SCNN \cite{pan2018spatial} formulates lane detection as multi-class semantic segmentation and is the basis of the 1st-place solution in TuSimple challenge \cite{tusimple}. It's core spatial CNN module recurrently aggregates spatial information to complete the discontinuous segmentation predictions, which then requires heuristic post-processing to decode the segmentation map. Hence, it has a high latency, and only struggles to be real-time after an optimization of Zheng~\etal \cite{zheng2021resa}. Others explore knowledge distillation \cite{hou2019learning} or generative modeling \cite{ghafoorian2018gan}, but their performance merely surpasses the seminal SCNN. Moreover, these methods typically assume a fixed number (\eg, 4) of lines. LaneNet~\cite{neven2018towards} leverages an instance segmentation pipeline to deal with a variable number of lines, but it requires post-inference clustering to generate line instances.
Some methods leverage row-wise classification~\cite{yoo2020end,qin2020ultra}, which is a customized down-sampling of per-pixel segmentation so that they still require post-processing. 
Qin \etal \cite{qin2020ultra} propose to trade performance for low latency, but their use of fully-connected layers results in large model size.

In short, segmentation-based methods all require heavy post-processing due to the misalignment of representations. They also suffer from the locality of segmentation task, so that they tend to perform worse under occlusions or extreme lighting conditions.%

\noindent \textbf{Point Detection-based Lane Detection.}
The success of object detection methods drives researchers to formulate lane detection as to detect lanes as a series of points (\eg, every 10 pixels in the vertical axis). %
Line-CNN \cite{li2019line} adapts classic Faster R-CNN \cite{ren2015faster} as a one-stage lane line detector, but it has a low inference speed (\textless $30$ FPS). Later, LaneATT \cite{tabelini2021keep} adopts a more general one-stage detection approach that achieves superior performance. 

However, these methods have to design heuristic lane anchors, which highly depend on dataset statistics, and require the Non-Maximum Suppression (NMS) as post-processing. On the contrary, we represent lane lines as curves with a fully end-to-end pipeline (anchor-free, NMS-free).

\noindent \textbf{Curve-based Lane Detection.} The pioneering work~\cite{van2019end} proposes a differentiable least squares fitting module to fit a polynomial curve (\eg, $x = ay^3 + by^2 + cy + d$) to points predicted by a deep neural network. The PolyLaneNet \cite{tabelini2021polylanenet} then directly learns to predict the polynomial coefficients with simple fully-connected layers. Recently, LSTR \cite{liu2021end} uses transformer blocks to predict polynomials in an end-to-end fasion based on the DETR \cite{carion2020end}.

Curve is a holistic representation of lane line, which naturally eliminates occlusions, requires no post-processing, and can predict a variable number of lines. 
However, their performance on large and challenging datasets (\eg, CULane \cite{pan2018spatial} and LLAMAS \cite{llamas2019}) still lag behind methods of other categories. They also suffer from slow convergence (over $2000$ training epochs on TuSimple), high latency architecture (\eg, LSTR \cite{liu2021end} uses transformer blocks which are difficult to optimize for low latency). 
We attribute their failure to the difficult-to-optimize and abstract polynomial coefficients. We propose to use the parametric Bézier curve, which is defined by actual control points on the image coordinate system\footnote{Actually control points of Bézier curves can be outside the image, but statistically that rarely happens in autonomous driving scenes.}, to address these problems.%

\begin{table}[t]
    \centering
    \resizebox{4cm}{!}{\begin{tabular}{CCC}
    \toprule
        n & \textbf{Bézier} & \textbf{Polynomial} \\
        \toprule
        2\mathrm{nd} & \mathbf{0.653} & 0.945 \\
        3\mathrm{rd} & \mathbf{0.471} & 0.558 \\
        4\mathrm{th} & \mathbf{0.315} & 0.330 \\
        \bottomrule
    \end{tabular}}
    \caption{Comparison of $n$-order Bézier curves and polynomials ($x = \sum_{i=0}^{n} a_{i} y^i $) on TuSimple \cite{tusimple} \textit{test} set (\textbf{lower is better}). Since the official metrics are too lose to show any meaningful difference, we use the fine-grained LPD metric following \cite{tabelini2021polylanenet}.}
    \label{tab:gtcompare}
    \vspace{-3mm}
\end{table}

\noindent \textbf{Bézier curve in Deep Learning.} To our knowledge, the only known successful application of Bézier curves in deep learning is the ABCNet~\cite{liu2020abcnet}, which uses cubic Bézier curve for text spotting. 
However, their method cannot be directly used for our tasks.
First, this method still uses NMS so that it cannot be end-to-end. We show in our work that NMS is not necessary so that our method can be an end-to-end solution.
Second, it calculates $L_1$ loss directly on the sparse Bézier control points, which results in difficulties of optimization.
We address this problem in our work by leveraging a fine-grained sampling loss.
In addition, we propose the feature flip fusion module, which is specifically designed for the lane detection task.

\section{BézierLaneNet}
\label{sec:BézierLaneNet}

\subsection{Overview}
\label{sec:over}

\noindent \textbf{Preliminaries on Bézier Curve.} The Bézier curve's formulation is shown in~\Cref{eq:bezier}, which is a parametric curve defined by $n+1$ control points:

\vspace{-3mm}
\begin{align}
\label{eq:bezier}
    \mathcal{B}(t) = \sum_{i=0}^{n} b_{i,n}(t) \mathcal{P}_{i},~0 \leq t \leq 1,
\end{align}
where $\mathcal{P}_{i}$ is the $i-th$ control point, $b_{i,n}$ are Bernstein basis polynomials of degree $n$:
\begin{align}
\label{eq:bernstein}
    b_{i,n} = C_{n}^{i} t^{i}(1 - t)^{n - i},~i=0,...,n.
\end{align}

We use the classic cubic Bézier curve ($n=3$), which is empirically found sufficient for modeling lane lines. It shows better ground truth fitting ability than $3$rd order polynomial (\Cref{tab:gtcompare}), which is the base function for previous curve-based methods \cite{tabelini2021polylanenet,liu2021end}. Higher-order curves do not bring substantial gains while the high degrees of freedom leads to instability. All coordinates for points discussed here are relative to the image size (\ie, mostly in range $[0,1]$). %

\begin{figure}[t]
    \centering
    \includegraphics[scale=0.5]{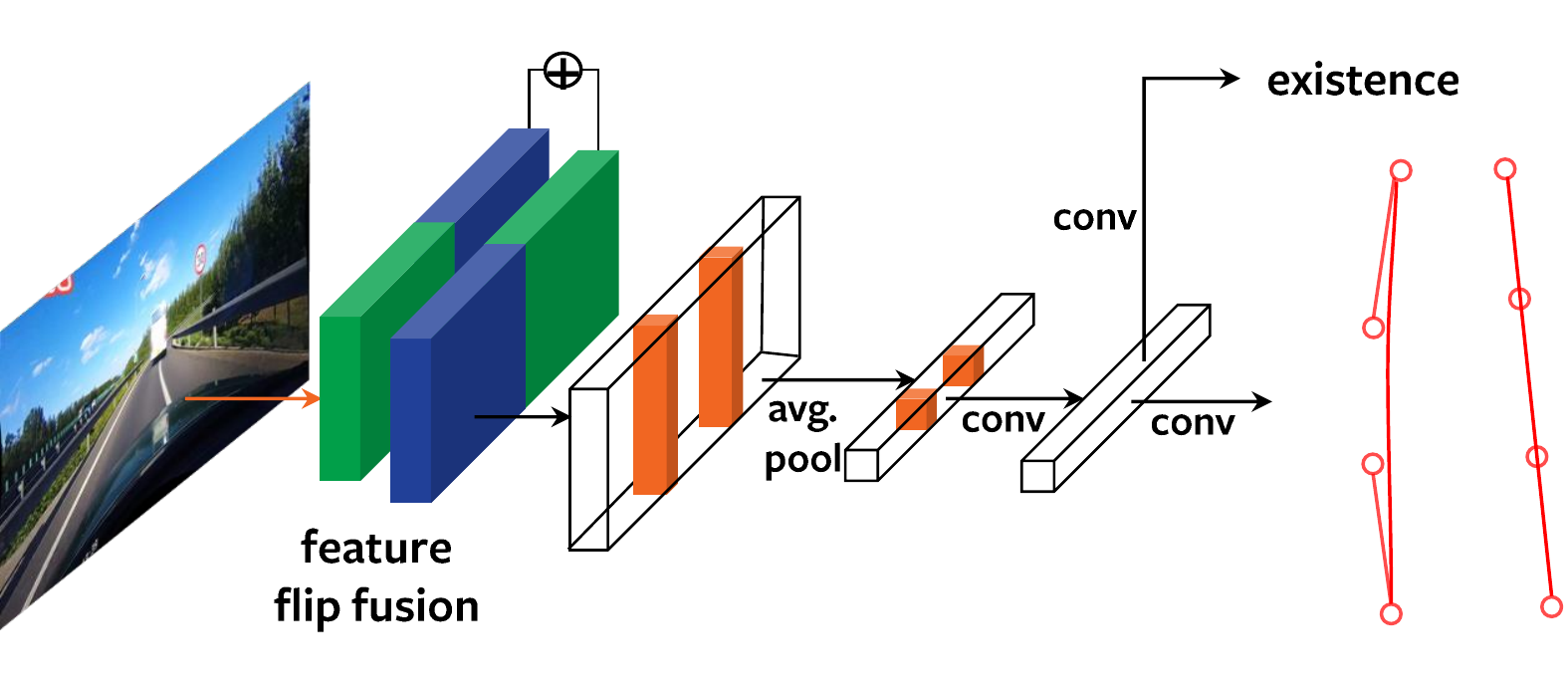}
    \vspace{-1mm}
    \caption{Pipeline. Feature from a typical encoder (\eg, ResNet) is strengthened by feature flip fusion, then pooled to 1D and two 1D convolution layers are applied. At last the network predicts Bézier curves through a classification branch and a regression branch.}
    \label{fig:pipeline}
    \vspace{-2mm}
\end{figure}

\noindent \textbf{The Proposed Architecture.} The overall model architecture is shown in \Cref{fig:pipeline}. Specifically, we use layer-$3$ feature of ResNets \cite{he2016deep} as backbone following RESA \cite{zheng2021resa}, but we replace the dilation inside the backbone network by two dilated blocks outside with dilation rates $[4, 8]$ \cite{chen2021you}. This strikes a better speed-accuracy trade-off for our method, which leaves a $16\times$ down-sampled feature map with a larger receptive field. We then add the feature flip fusion module (\Cref{sec:feature}) to aggregate opposite lane features.
The enriched feature map ($C \times \frac{H}{16} \times \frac{W}{16}$) is then pooled to ($C \times \frac{W}{16}$) by average pooling, resulting in $\frac{W}{16}$ proposals ($50$ for CULane \cite{pan2018spatial}). 
Two $1\times3$ 1D convolutions are used to transform the pooled features, while also conveniently modeling interactions between nearby lane proposals, guiding the network to learn a substitute for the non-maximal suppression (NMS) function. 
Lastly, the final prediction is obtained by the classification and regression branches (each is only one $1\times1$ 1D convolution). The outputs are $\frac{W}{16} \times 8$ for regression of 4 control points, and $\frac{W}{16} \times 1$ for existence of lane line object.

\subsection{Feature Flip Fusion}
\label{sec:feature}

By modeling lane lines as holistic curves, we focus on the geometric properties of individual lane lines (\eg, thin, long, and continuous). Now we consider the global structure of lanes from a front-mounted camera view in driving scenes.
Roads have equally spaced lane lines, which appear symmetrical and this property is worth modeling.
For instance, the existence of left ego lane line should very likely indicate its right counterpart, the structure of immediate left lane could help describe the immediate right lane, \etc.

\begin{figure}[t]

\centering
    \begin{tabular}{c}
    \includegraphics[width = 0.99\linewidth, height=3.2cm]{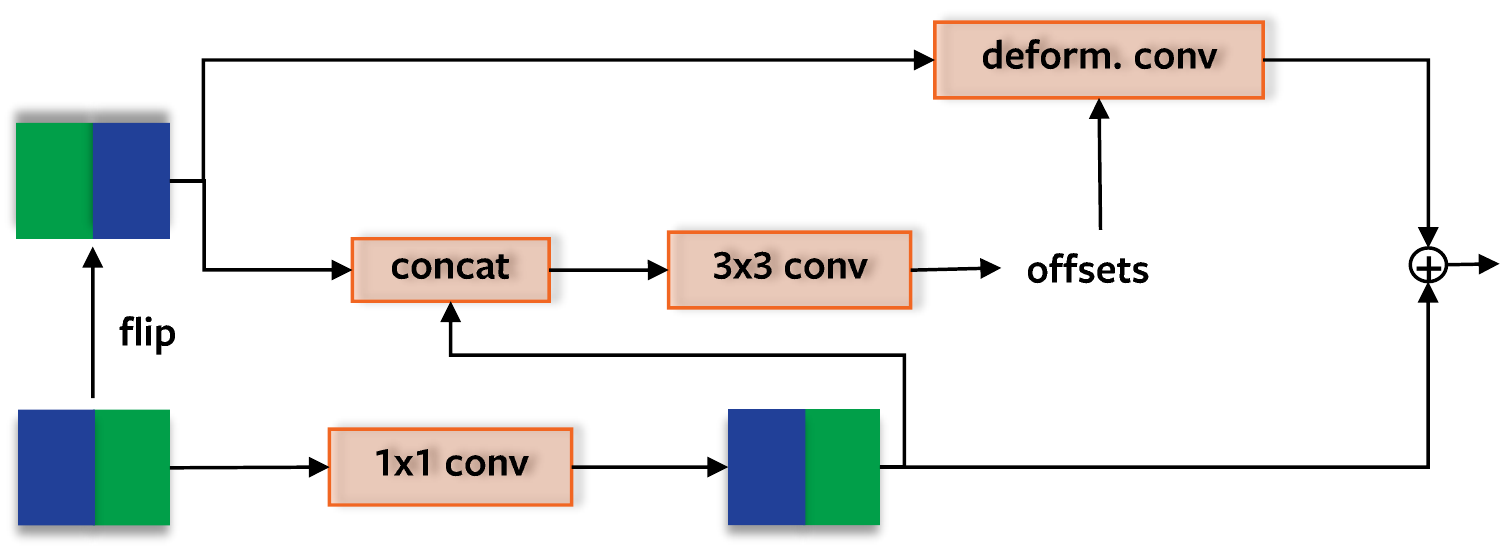}\\
    \end{tabular}
    \caption{Feature flip fusion. Alignment is achieved by calculating deformable convolution offsets, conditioned on both the flipped and original feature map. Best viewed in color.}
\vspace{-4mm}
\label{fig:flip}
\end{figure}

To exploit this property, we fuse the feature map with its horizontally flipped version (\Cref{fig:flip}). Specifically, two separate convolution and normalization layers transform each feature map, they are then added together before a ReLU activation. With this module, we expect the model to base its predictions on both feature maps.

To account for the slight misalignment of camera captured image (\eg, rotated, turning, non-paired), we apply deformable convolution \cite{zhu2019deformable} with kernel size $3\times3$ for the flipped feature map while learning the offsets conditioned on the original feature map for feature alignment. 

We add an auxiliary binary segmentation branch (to segment lane line or non-lane line areas, which would be removed after training) to the ResNet backbone, and we expect it to enforce the learning of spatial details.
Interestingly, we find this auxiliary branch improves the performance only when it works with the feature fusion. This is because the localization of the segmentation task may provide a more spatially-accurate feature map, which in turn supports accurate fusion between the flipped features.

Visualizations are shown in \Cref{fig:featurevis}, from which we can see that the flipped feature does correct the error caused by the asymmetry introduced by the car (\Cref{fig:featurevis}(a)).

\subsection{End-to-end Fit of a Bézier Curve}
\label{sec:bezierloss}

\noindent \textbf{Distances Between Bézier Curves.} The key to learning Bézier curves is to define a good distance metric measuring the distances between the ground truth curve and prediction.
Naively, one can directly calculate the mean $L_1$ distance between Bézier curve control points, as in ABCNet \cite{liu2020abcnet}. However, as shown in \Cref{fig:loss}(a), a large $L_1$ error in curvature control points can demonstrate a very small visual distance between Bézier curves, especially on small or medium curvatures (which is often the case for lane lines). Since Bézier curves are parameterized by $t \in [0, 1]$, we propose the more reasonable sampling loss for Bézier curves (\Cref{fig:loss}(b)), by sampling curves at a uniformly spaced set of $t$ values ($T$), which means equal curve length between adjacent sample points. %
The $t$ values can be further transformed by a re-parameterization function $f(t)$. Specifically, given Bézier curves $\mathcal{B}(t), \mathcal{\hat{B}}(t)$, the sampling loss $\mathcal{L}_{reg}$ is:

\vspace{-5mm}
\begin{align}
    \mathcal{L}_{reg} = \frac{1}{n} \sum_{t \in T} ||\mathcal{B}(f(t)) - \mathcal{\hat{B}}(f(t))||_{1},
\label{eq:losssample}
\vspace{-2mm}
\end{align}
where $n$ is the total number of sampled points and is set to $100$. We empirically find $f(t) = t$ works well.
This simple yet effective loss formulation makes our model easy to converge and less sensitive to hyper-parameters that typically involved in other curved-based or point detection-based methods, \eg, loss weighting for endpoints loss \cite{liu2021end} and line length loss \cite{tabelini2021keep} (see \Cref{fig:loss}(b,c)).

\def\imh{2.8cm}
\begin{figure}[t]
\renewcommand{\tabcolsep}{1pt}
\centering
    \begin{tabular}{c}
    \includegraphics[width = 0.95\linewidth, height=\imh]{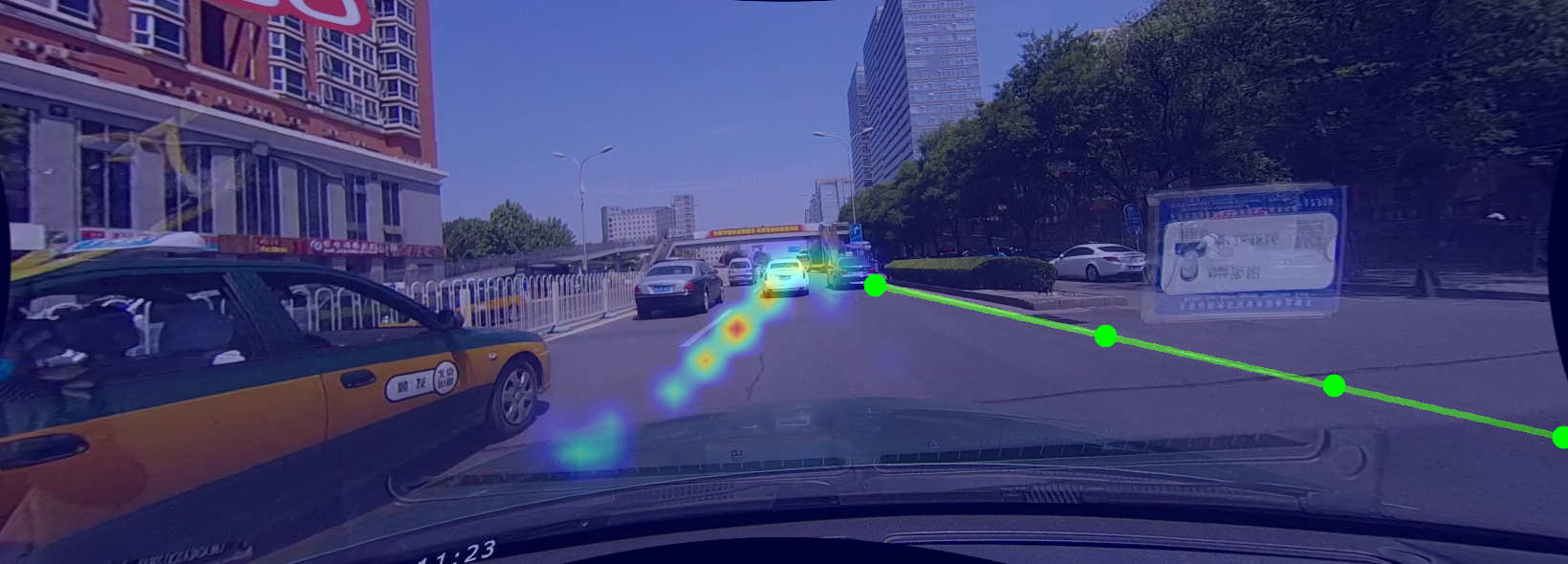} \\
    (a) \\
    \includegraphics[width = 0.95\linewidth, height=\imh]{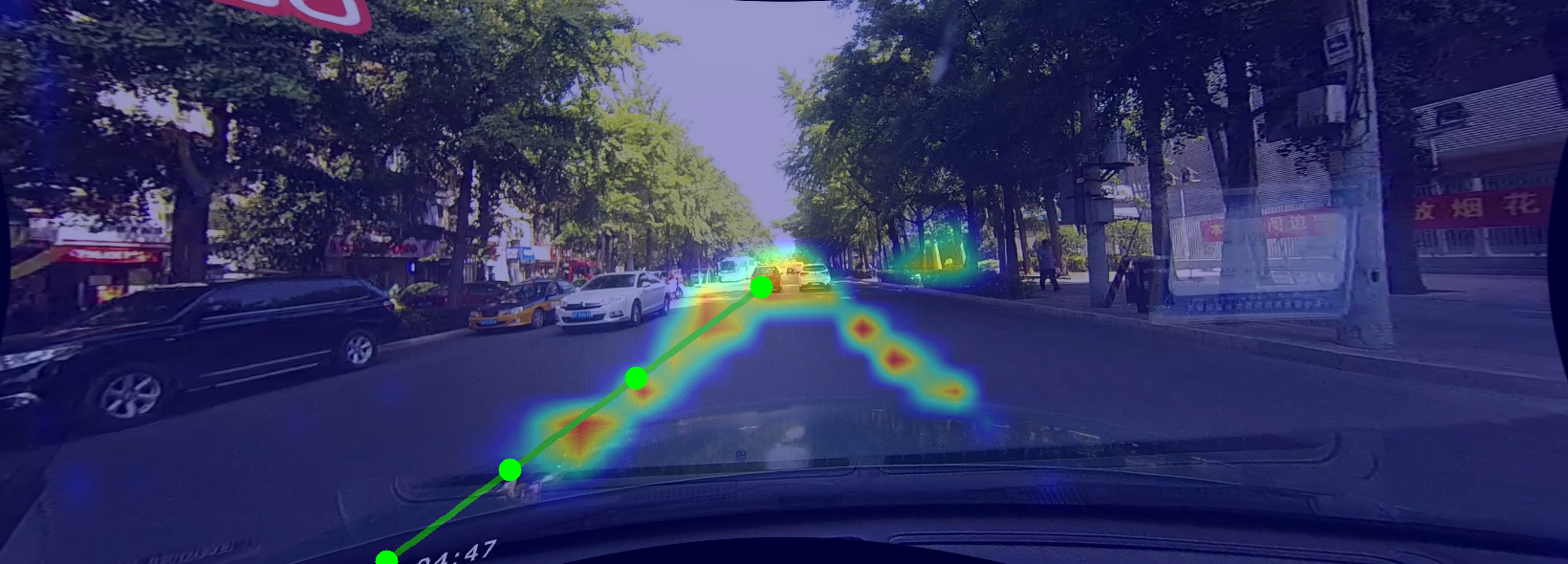} \\
    (b) \\
    \end{tabular}
    \vspace{-2mm}
\caption{Grad-CAM \cite{selvaraju2017grad} visualization on the last layer of ResNet backbone. (a) Our model can infer existence of an ill-marked lane line, from clear markings and cars around the opposite line. Note that the car is deviated to the left, this scene was not captured with perfect symmetry. (b) When entire road lacks clear marking, both sides are used for a better prediction. Best viewed in color.}
\label{fig:featurevis}
\vspace{-4mm}
\end{figure}

\noindent \textbf{Bézier Ground Truth Generation.} Since lane datasets are currently annotated by on-line key points, we need the Bézier control points for the above sampling loss.
Given the annotated points $\{(k_{x_i}, k_{y_i})\}_{i=1}^m$ on one lane line, where $(k_{x_i}, k_{y_i})$ denotes the 2D-coordinates of the $i$-th point.
Our goal is to obtain control points $\{\mathcal{P}_i({x_i}, {y_i})\}_{i=1}^n$.
Similar to~\cite{liu2020abcnet}, we use least squares fitting to solve this equation:
\vspace{-2mm}
\begin{equation}
\renewcommand\arraycolsep{1.5pt}
\begin{bmatrix}
  b_{0,n}(t_0) & \cdots\ & b_{n,n}(t_0)\\
  b_{0,n}(t_1) & \cdots\ & b_{n,n}(t_1)\\
 \vdots&  \ddots\ & \vdots\\
  b_{0,n}(t_m) & \cdots\ & b_{n,n}(t_m)
\end{bmatrix}
\begin{bmatrix}
 \mathcal{P}_{0}\\
 \mathcal{P}_{1}\\
 \vdots \\
 \mathcal{P}_{n}\\
\end{bmatrix}
=
\begin{bmatrix}
 k_{x_0} & k_{y_0}\\
 k_{x_1} & k_{y_1}\\
 \vdots & \vdots\\
 k_{x_m} & k_{y_m}\\
\end{bmatrix},
\label{eq:matrix}
\end{equation}
$\{t_i\}_{i=0}^m \in [0, 1]$. Considering $m >> n$ and the linear independence of $b$, it can be efficiently solved by a pseudo-inverse of the leftmost matrix (mostly full column rank). Different from \cite{liu2020abcnet}, we do not set $\mathcal{P}_{0}$ and $\mathcal{P}_{n}$ to original endpoints, which leads to better quality labels.

\def\imh{1.4in}
\begin{figure}[t]
\renewcommand{\tabcolsep}{1pt}
\centering
    \begin{tabular}{ccc}
    \includegraphics[width = 0.33\linewidth, height=\imh]{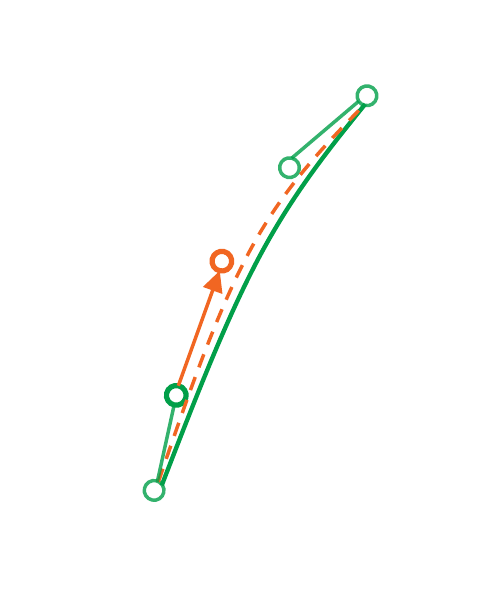}&
    \includegraphics[width = 0.33\linewidth, height=\imh]{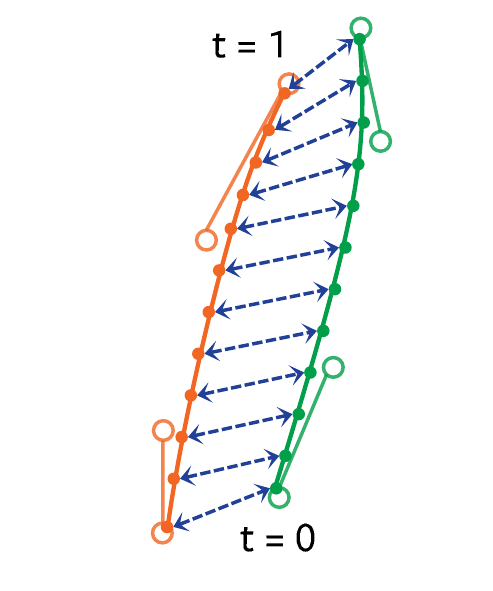}&
    \includegraphics[width = 0.33\linewidth, height=\imh]{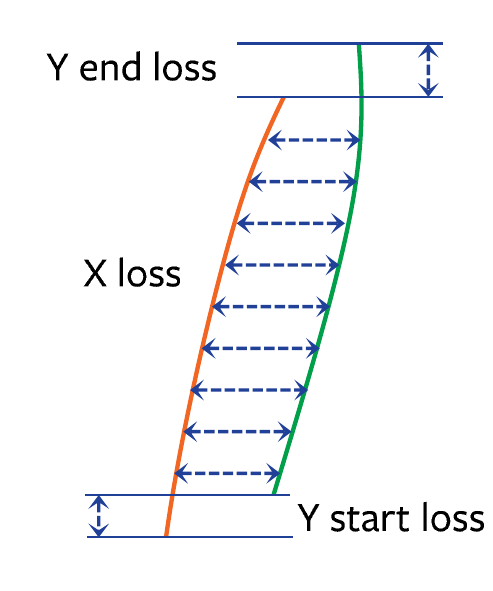}\\
    (a) & (b) & (c) \\
    \end{tabular}
\caption{Lane loss functions. (a) The $L_1$ distance of control points is not highly correlated with the actual distance between curves. (b) The proposed sampling loss is one unified distance metric by $t$-sampling. (c) Typical loss for polynomial regression \cite{liu2021end}, at least 3 separate losses are required: $y$-sampling loss, $y$ start point loss, $y$ end point loss.}
\label{fig:loss}
\end{figure}

\noindent \textbf{Label and Prediction Matching.} After obtaining the ground truth, in training, we perform a one-to-one assignment between $G$ labels and $N$ predictions ($G < N$) using optimal bipartite matching, to attain a fully end-to-end pipeline. Following Wang \etal \cite{wang2021end}, we find a $G$-permutation of $N$ predictions $\pi \in \Pi_G^N$ that formulates the best bipartite matching:

\vspace{-6mm}

\begin{align}
    \hat{\pi} = \argmax_{\pi \in \Pi_G^N} \sum_i^G Q_{i,\pi(i)},
\label{eq:bipartite}
\end{align}
\vspace{-5mm}
\begin{align}
    Q_{i,\pi(i)} = {\Bigl( \hat{p}_{\pi(i)} \Bigr)}^{1-\alpha} ~\cdot {\Bigl(1 - L_1 \bigl( b_i, \hat{b}_{\pi(i)} \bigr)\Bigr)}^\alpha,
\label{eq:quality}
\end{align}
where $Q_{i,\pi(i)} \in [0,1]$ represents matching quality of the $i$-th label with the $\pi(i)$-th prediction, based on $L_1$ distance between curves $b_i, \hat{b}_{\pi(i)}$ (sampling loss) and class score $\hat{p}_{\pi(i)}$. $\alpha$ is set to $0.8$ by default. The above equations can be efficiently solved by the well-known Hungarian algorithm.

Wang \etal \cite{wang2021end} also use a spatial prior that restricts the matched prediction to a spatial neighborhood of the label (object center distance, the \textit{centerness} prior in FCOS \cite{tian2019fcos}). However, since lots of lanes are long lines with a large slope, this centerness prior is not useful. See \Cref{sec:match} for more investigations on matching priors.

\noindent \textbf{Overall Loss.} Other than Bézier curve sampling loss, there is also the classification loss $\mathcal{L}_c$ for the lane object classification (existence) branch. Since the imbalance between positive and negative examples is not as severe in lane detection as in object detection, instead of the focal loss \cite{lin2017focal}, we use the simple weighted binary cross-entropy loss:

\vspace{-5mm}

\begin{align}
    \mathcal{L}_{cls} = -{(y\log(p) + w(1 - y)\log(1 - p))},
\label{eq:lossclass}
\end{align}
where $w$ is the weighting for negative samples, which is set to $0.4$ in all experiments.
The loss $\mathcal{L}_{seg}$ for the binary segmentation branch (\Cref{sec:feature}) takes the same format.

The overall loss is a weighted sum of all three losses:

\vspace{-5mm}

\begin{align}
    \mathcal{L} = \lambda_{1}\mathcal{L}_{reg} + \lambda_{2}\mathcal{L}_{cls} + \lambda_{3}\mathcal{L}_{seg},
\label{eq:lossall}
\end{align}
where $\lambda_1, \lambda_2, \lambda_3$ are set to $1, 0.1, 0.75$, respectively.

\section{Experiments}
\label{sec:exper}

\subsection{Datasets}
\label{sec:dataset}
To evaluate the proposed method, we conduct experiments on three well-known datasets: TuSimple ~\cite{tusimple}, CULane ~\cite{pan2018spatial} and LLAMAS \cite{llamas2019}.
TuSimple dataset was collected on highways with high-quality images, under fair weather conditions.
CULane dataset contains more complex urban driving scenarios, including shades, extreme illuminations, and road congestion.
LLAMAS is a newly formed large-scale dataset, it is the only lane detection benchmark without public \textit{test} set labels.
Details of these datasets can be found in \Cref{tab: datasets}.

\begin{table}[t]
\begin{center}
\scalebox{0.82}{
\begin{tabular}{lCCCCC}
\toprule
\textbf{Dataset}    & \textbf{Train} & \textbf{Val} & \textbf{Test} & \textbf{Resolution} & \# \textbf{Lines} \\
\toprule
TuSimple \cite{tusimple} & 3268 & 358 & 2782 & 720 \times 1280 & \le5 \\
CULane \cite{pan2018spatial} & 88880 & 9675 & 34680 & 590 \times 1640  & \le4 \\
LLAMAS \cite{llamas2019} & 58269 & 20844 & 20929 & 717 \times 1276 & \le4^{*} \\
\bottomrule
\end{tabular}}
\end{center}
\vspace{-5mm}
\caption{Details of datasets. *Number of lines in LLAMAS dataset is more than 4, but official metric only evaluates 4 lines.}
\label{tab: datasets}
\vspace{-3mm}
\end{table}

\subsection{Evalutaion Metics}
For CULane \cite{pan2018spatial} and LLAMAS \cite{llamas2019}, the official metric is F1 score from \cite{pan2018spatial}:
\begin{align}
\label{eq:f1}
    \mathrm{F1} = \frac{2 \cdot \mathrm{Precision} \cdot \mathrm{Recall}}{\mathrm{Precision} + \mathrm{Recall}},
\end{align}
where $\mathrm{Precision} = \frac{TP}{TP + FP}$ and $\mathrm{Recall} = \frac{TP}{TP + FN}$. Lines are assumed to be $30$ pixels wide, prediction and ground truth lines with pixel IoU over $0.5$ are considered a match.

For TuSimple~\cite{tusimple} dataset, the official metrics include Accuracy, false positive rate (FPR), and false negative rate (FNR). 
Accuracy is computed as $\frac{N_{pred}}{N_{gt}}$,
where $N_{pred}$ is the number of correctly predicted on-line points and $N_{gt}$ is the number of ground truth on-line points.

\subsection{Implementation Details}
\label{sec:imple}

\begin{table*}[t]
    \centering
    \resizebox{16cm}{!}{
    \begin{tabular}{lCCCCCCCCCCCcCCCC}
        \toprule
    &\multicolumn{11}{c}{\textbf{CULane \cite{pan2018spatial}}} & \multicolumn{5}{c}{\textbf{TuSimple \cite{tusimple}}}  \\
    \cmidrule(lr){2-12}
    \cmidrule(lr){13-17}
        \textbf{Method} & \textbf{Ep.} & \textbf{Total} & \textbf{Normal} & \textbf{Crowd} & \textbf{Night} & \makecell{\textbf{No} \\ \textbf{line}} & \textbf{Shadow} & \textbf{Arrow} & \makecell{\textbf{Dazzle} \\ \textbf{light}} & \textbf{Curve} & \textbf{Cross $\downarrow$} & train+val & \textbf{Ep.} & \textbf{Acc.} & \textbf{FPR $\downarrow$} & \textbf{FNR $\downarrow$} \\
        \toprule
        \textbf{Segmentation-based} \\
        \cmidrule{1-1}
        Baseline (ResNet-18)* & 12 & 65.30 & 85.45 & 62.63 & 61.04 & 33.88 & 51.72 & 78.15 & 53.05 & 59.70 & 1915 & & 50 & 94.25 & 0.088 & 0.089 \\
        Baseline (ResNet-34)* & 12 & 69.92 & 89.46 & 66.66 & 65.38 & 40.43 & 62.17 & 83.18 & 58.51 & 63.00 & 1713 & & 50 & 95.31 & 0.064 & 0.062 \\
        Baseline (ResNet-101)* & 12 & 71.37 & 90.11 & 67.89 & 67.01 & 43.10 & 70.56 & 85.09	& 61.77 & 65.47 & 1883 & & 50 & 95.19 & 0.062 & 0.062 \\
        SCNN (ResNet-18) \cite{pan2018spatial}* & 12 & 72.19 & 90.98 & 70.17 & 66.54 & 43.12 & 66.31 & 85.62 & 62.20 & 65.58 & 1808 & & 50 & 94.77 & 0.075 & 0.074\\
        SCNN (ResNet-34) \cite{pan2018spatial}* & 12 & 72.70 & 91.06	& 70.41	& 67.75 & 44.64	& 68.98	& 86.50	& 61.57	& 65.75	& 2017 & & 50 & 95.25 & 0.063 & 0.063  \\
        SCNN (ResNet-101) \cite{pan2018spatial}* & 12 & 73.58 & 91.10 & 71.43 & 68.53 & 46.39 & 72.61 & 86.87 & 61.95 & 67.01 & 1720 & & 50 & \boldred{95.69} & \red{0.052} & \red{0.050} \\
        UFLD (ResNet-18) \cite{qin2020ultra}** & 50 & 68.4 & 87.7 & 66.0 & 62.1 & 40.2 & 62.8 & 81.0 & 58.4 & 57.9 & 1743 & - & - & - & - & -  \\
        UFLD (ResNet-34) \cite{qin2020ultra}** & 50 & 72.3 & 90.7 & 70.2 & 66.7 & 44.4 & 69.3 & 85.7 & 59.5 & \red{69.5} & 2037 & - & - & - & - & -  \\
        RESA (ResNet-18) \cite{zheng2021resa}* & 12 & 72.90 & 91.23 & 70.57 & 67.16 & 45.24 & 68.01 & 86.56 & 64.32 & 66.19 & 1679 & & 50 & 95.24 & 0.069 & 0.057  \\
        RESA (ResNet-34) \cite{zheng2021resa}* & 12 & 73.66 & 91.31 & \red{71.80} & 67.54 & \red{46.57}	& 72.74	& 86.94	& \red{64.46}	& 67.31	& 1701 & & 50 & 95.15 & 0.069 & 0.059 \\
        RESA (ResNet-101) \cite{zheng2021resa}* & 12 & \boldred{74.04} & \red{91.45} & 71.51 & \red{69.01} & 46.54 & \red{75.83} & \red{87.75} & 63.90 & 68.24 & \red{1522} & & 50 & 95.56 & 0.058 & 0.051 \\
        \midrule
        \textbf{Point detection-based} \\
        \cmidrule{1-1}
        FastDraw (ResNet-18) \cite{philion2019fastdraw} & - & - & - & - & - & - & - & - & - & - & - & \checkmark & 7 & 94.9 & 0.061 & 0.047 \\
        CurveLanes-NAS-S \cite{xu2020curvelane} & 12 & 71.4 & 88.3 & 68.6 & 66.2 & 47.9 & 68.0 & 82.5 & 63.2 & 66.0 & 2817 & - & - & - & - & -  \\
        CurveLanes-NAS-M \cite{xu2020curvelane} & 12 & 73.5 & 90.2 & 70.5 & 68.2 & 48.8 & 69.3 & 85.7 & 65.9 & 67.5 & 2359 & - & - & - & - & - \\
        CurveLanes-NAS-L \cite{xu2020curvelane} & 12 & 74.8 & 90.7 & 72.3 & 68.9 & 49.4 & 70.1 & 85.8 & 67.7 & \red{68.4} & 1746 & - & - & - & - & - \\
        LaneATT (ResNet-18) \cite{tabelini2021keep}** & 15 & 74.88 & 90.98 & 72.78 & 68.61 & 48.23 & 69.68 & 85.44 & 65.43 & 63.18 & \red{1163} & \checkmark & 100 & 95.57 &0.036  & 0.030 \\
        LaneATT (ResNet-34) \cite{tabelini2021keep}** & 15 & 76.42 & \red{91.94}                             & 74.76               & 70.32           & 49.17                                   & \red{77.68}                       & \red{88.14}       & 65.92                               & 68.07                             & 1323                           & \checkmark & 100 & 95.63 & \red{0.035} & 0.029  \\
        LaneATT (ResNet-122) \cite{tabelini2021keep}** & 15 & \boldred{76.79} & 91.50 & \red{76.04} & \red{70.43} & \red{50.29} & 75.96 & 86.16 & \red{68.99} & 63.99 & 1265 & \checkmark & 100 & \boldred{96.10} & 0.056 & \red{0.022} \\
        \midrule
        \textbf{Curve-based} \\
        \cmidrule{1-1}
        PolyLaneNet (EfficientNet-B0) \cite{tabelini2021polylanenet}** & - & - & - & - & - & - & - & - & - & - & - & \checkmark & 2695 & 93.36 & 0.094 & 0.093  \\
        LSTR (ResNet-18, $1\times$) \cite{liu2021end}* & - & - & - &- &- &- &- &- &- &- & - &  & 2000 & 95.06 & \red{0.049} & 0.042 \\
        LSTR (ResNet-18, $2\times$) \cite{liu2021end}* & 150 & 68.72 & 86.78 &67.34 &59.92 &40.10 &59.82 &78.66 &56.63 &56.64 & 1166 & - & - & - & - & - \\
        \textbf{BézierLaneNet (ResNet-18)} & 36 & 73.67 & 90.22 &71.55 &68.70 &45.30 &70.91 &84.09 &62.49 &58.98 &996 & & 400 & 95.41 & 0.053 & 0.046 \\
        \textbf{BézierLaneNet (ResNet-34)} & 36 & \boldred{75.57} &  \red{91.59} & \red{73.20}  & \red{69.90} & \red{48.05} & \red{76.74} & \red{87.16} & \red{69.20} & \red{62.45} & \red{888} & & 400 & \boldred{95.65} & 0.051 & \red{0.039} \\
        \bottomrule
    \end{tabular}}
    \caption{Results on \textit{test} set of CULane \cite{pan2018spatial} and TuSimple \cite{tusimple}. *reproduced results in our code framework, best performance from three random runs. **reported from reliable open-source codes from the authors.}
    \label{tab:all}
    \vspace{-2mm}
\end{table*}

\noindent \textbf{Fair Comparison.} To fairly compare among different state-of-the-art methods, we re-implement representative methods \cite{pan2018spatial,zheng2021resa,liu2021end} in a unified PyTorch framework. We Also provide a semantic segmentation baseline \cite{deeplabv1} originally proposed in \cite{pan2018spatial}. All our implementations do \textbf{not} use \textit{val} set in training, and tune hyper-parameters \textbf{only} on \textit{val} set. Some methods with reliable open-source codes are reported from their own codes \cite{qin2020ultra,tabelini2021keep,tabelini2021polylanenet}. For platform sensitive metric Frames-Per-Second (FPS), we re-evaluate all reported methods on the same RTX 2080 Ti platform. More details for implementations and FPS tests are in \Cref{sec:fps,sec:specs,sec:transforms}.

\noindent \textbf{Training.} We train 400, 36, 20 epochs for TuSimple, CULane, and LLAMAS, respectively (training of our model takes only $12$ GPU hours on a single RTX 2080 Ti), and the input resolution is $288 \times 800$ for CULane \cite{pan2018spatial} and $360 \times 640$ for others, following common practice.  Other than these, all hyper-parameters are tuned on CULane \cite{pan2018spatial} \textit{val} set and remain the same for our method across datasets. We use Adam optimizer with learning rate $6\times10^{-4}$, weight decay $1\times10^{-4}$, batch size 20, Cosine Annealing learning rate schedule as in \cite{tabelini2021keep}. Data augmentation includes random affine transforms, random horizontal flip, and color jitter.

\noindent \textbf{Testing.} No post-processing is required for curve methods.
Standard Gaussian blur and row selection post-processing is applied to segmentation methods. NMS is used for LaneATT \cite{tabelini2021keep}, while we remove its post-inference B-Spline interpolation in CULane \cite{pan2018spatial}, to align with our framework.

\subsection{Comparisons}
\label{sec:compar}

\noindent \textbf{Overview.} Experimental results are shown in \Cref{tab:all,tab:llamas}. TuSimple \cite{tusimple} is a small dataset that features clear-weather highway scenes and has a relatively easy metric, most methods thrive in this dataset. Thus, we mainly focus on the other two large-scale datasets \cite{pan2018spatial,llamas2019}, where there is still a rather clear difference between methods. For high-performance methods ($>70\%$ F1 on CULane \cite{pan2018spatial}), we also show efficiency metrics (FPS, Parameter count) in \Cref{tab:profile}.

\noindent \textbf{Comparison with Curve-based Methods.} As shown in \Cref{tab:all,tab:llamas}, in all datasets, BézierLaneNet outperforms previous curve-based methods \cite{liu2021end,tabelini2021polylanenet} by a clear margin, advances the state-of-the-art of curve-based methods by $6.85\%$ on CULane \cite{pan2018spatial} and $6.77\%$ on LLAMAS \cite{llamas2019}. Thanks to our fully convolutional and fully end-to-end pipeline, BézierLaneNet runs over $2\times$ faster than LSTR \cite{liu2021end}. LSTR has a speed bottleneck from transformer architecture, the $1\times$ and $2\times$ model have FPS 98 and 97, respectively\footnote{The original 420 FPS report from LSTR paper \cite{liu2021end}, is throughput with batch size 16, detailed discussions in \Cref{sec:fps}.}. While curves are difficult to learn, our method converges $4$-$5\times$ faster than LSTR. For the first time, an elegant curve-based method can challenge well-designed segmentation methods or point detection methods on these datasets while showing a favorable trade-off, with an acceptable convergence time.

\noindent \textbf{Comparison with Segmentation-based Methods.} These methods tend to have a low speed due to recurrent feature aggregation \cite{pan2018spatial,zheng2021resa}, and the use of high-resolution feature map \cite{deeplabv1,pan2018spatial,zheng2021resa}. BézierLaneNet outperforms them in both speed and accuracy. Our small models even compare favorably against RESA \cite{zheng2021resa} and SCNN \cite{pan2018spatial} with large ResNet-101 backbone, surpassing them in CULane \cite{pan2018spatial} with a clear margin ($1\sim2\%$).
On LLAMAS \cite{llamas2019}, where the dataset restricts testing on $4$ center lines, the segmentation approach shows strong performance (\Cref{tab:llamas}). Nevertheless, our ResNet-34 model still outperforms SCNN by $0.92\%$.

UFLD \cite{qin2020ultra} reformulates segmentation to row-wise classification on a down-sampled feature map to achieve fast speed, at the cost of accuracy. Compared to us, UFLD (ResNet-34) is $0.9\%$ lower on CULane \textbf{Normal}, while $7.4\%$, $3.0\%$, $3.2\%$ worse on \textbf{Shadow}, \textbf{Crowd}, \textbf{Night}, respectively. Overall, our method with the same backbones outperforms UFLD by $3\sim5\%$, while being faster on ResNet-34. Besides, UFLD uses large fully-connected layers to optimize latency, which causes a huge model size (the largest in \Cref{tab:profile}).

A drawback for all segmentation methods is the weaker performance on \textbf{Dazzle Light}. Per-pixel (or per-pixel grid for UFLD \cite{qin2020ultra}) segmentation methods may rely on information from local textures, which is destroyed by extreme exposure to light. While our method predicts lane lines as holistic curves, hence robust to changes in local textures.

\begin{table}[t]
    \centering
    \resizebox{8cm}{!}{
    \begin{tabular}{lCCCC}
        \toprule
    &\multicolumn{4}{c}{\textbf{LLAMAS \cite{llamas2019}}}  \\
    \cmidrule(lr){2-5}
        \textbf{Method} & \textbf{Ep.} & \textbf{F1} & \textbf{Precision} & \textbf{Recall} \\
        \toprule
        \textbf{Segmentation-based} \\
        \cmidrule{1-1}
        Baseline (ResNet-34)* & 18 & 93.43 & 92.61 & 94.27 \\
        SCNN (ResNet-34) \cite{pan2018spatial}* & 18 & 94.25 & 94.11 & 94.39  \\
        \midrule
        \textbf{Point detection-based} \\
        \cmidrule{1-1}
        LaneATT (ResNet-18) \cite{tabelini2021keep}** & 15 & 93.46 & \red{96.92} & 90.24 \\
        LaneATT (ResNet-34) \cite{tabelini2021keep}** & 15 & 93.74 & 96.79 & 90.88  \\
        LaneATT (ResNet-122) \cite{tabelini2021keep}** & 15 & 93.54 & 96.82 & 90.47 \\
        \midrule
        \textbf{Curve-based} \\
        \cmidrule{1-1}
        PolyLaneNet (EfficientNet-B0) \cite{tabelini2021polylanenet}** & 75 & 88.40 & 88.87 & 87.93  \\
        \textbf{BézierLaneNet (ResNet-18)} & 20 & 94.91 & 95.71 & 94.13 \\
        \textbf{BézierLaneNet (ResNet-34)} & 20 & \boldred{95.17} & 95.89 & \red{94.46} \\
        \bottomrule
    \end{tabular}}
    \caption{Results from LLAMAS \cite{llamas2019} test server.}
    \label{tab:llamas}
    \vspace{-5mm}
\end{table}

\noindent \textbf{Comparison with Point Detection-based Methods.} Xu \etal \cite{xu2020curvelane} finds a series of point detection-based models with neural architecture search techniques called CurveLanes-NAS. Despite its complex pipeline and extensive architecture search for the best accuracy-FLOPs trade-off, our simple ResNet-34 backbone model ($29.9$ GFLOPs) still surpasses its large model ($86.5$ GFLOPs) by $0.8\%$ on CULane. CurveLanes-NAS also performs worse under occlusions, a similar drawback as the segmentation methods without recurrent feature fusion \cite{deeplabv1,qin2020ultra}. As shown in \Cref{tab:all}, with similar model capacity compared to our ResNet-34 model, CurveLanes-NAS-M ($35.7$ GFLOPs) is $1.4\%$ worse on \textbf{Normal} scenes, but the gap on \textbf{Shadow} and \textbf{Crowd} are $7.4\%$ and $2.7\%$.

Recently, LaneATT \cite{tabelini2021keep} achieves higher performance with a point detection network. However, their design is not fully end-to-end (requires Non-Maximal Suppression (NMS)), based on heuristic anchors (\textgreater $1000$), which are calculated directly from the dataset's statistics, thus may systematically pose difficulties in generalization. Still, with ResNet-34, our method outperforms LaneATT on the LLAMAS \cite{llamas2019} test server ($1.43\%$), with a significantly higher recall ($3.58\%$). We also achieve comparable performance to LaneATT on TuSimple \cite{tusimple} using only the \textit{train} set, and only $\sim1\%$ worse on CULane. Our method performs significantly better in \textbf{Dazzle Light} ($3.3\%$ better), comparably in \textbf{Night} ($0.4\%$ lower). It also has a lower False Positive (FP) rate on Crossroad scenes (\textbf{Cross}), even though LaneATT shows an extremely low-FP characteristic (large Precision-Recall gap in \Cref{tab:llamas}). Methods that rely on heuristic anchors \cite{tabelini2021keep} or heuristic decoding process \cite{pan2018spatial,zheng2021resa,qin2020ultra,xu2020curvelane} tend to have more false predictions in this scene. Moreover, the NMS is a sequential process that could have unstable runtime in real-world applications. Even when NMS was not evaluated on real inputs, our models are $29\%, 28\%$ faster, have $2.9\times$, $2.3\times$ fewer parameters, compared to LaneATT on ResNet-18 and ResNet-34 backbones, respectively. %

\begin{table}[t]
    \centering
    \resizebox{8cm}{!}{
    \begin{tabular}{lCC}
        \textbf{Method} &  \textbf{FPS $\uparrow$} & \textbf{Params (M) $\downarrow$}  \\
        \toprule
        \textbf{Segmentation-based (ignored post-processing time)} \\
        \cmidrule{1-1}
        Baseline (ResNet-101) & 27 & 43.56\\
        SCNN (ResNet-18) \cite{pan2018spatial} & 21 & 12.63 \\
        SCNN (ResNet-34) \cite{pan2018spatial} & 21 & 22.74 \\
        SCNN (ResNet-101) \cite{pan2018spatial} & 14 & 44.15 \\
        UFLD (ResNet-34) \cite{qin2020ultra} & 144 & 71.58 \\
        RESA (ResNet-18) \cite{zheng2021resa} & 68 & 6.61 \\
        RESA (ResNet-34) \cite{zheng2021resa} & 54 & 11.99 \\
        RESA (ResNet-101) \cite{zheng2021resa} & 25 & 31.46 \\
        \midrule
        \textbf{Point detection-based (ignored NMS time in real images)} \\
        \cmidrule{1-1}
        LaneATT (ResNet-18) \cite{tabelini2021keep} & 165 & 12.02 \\
        LaneATT (ResNet-34) \cite{tabelini2021keep}& 117 & 22.13 \\
        LaneATT (ResNet-122) \cite{tabelini2021keep}& 26 & 8.55 \\
        \midrule
        \textbf{Curve-based (entirely end-to-end)} \\
        \cmidrule{1-1}
        \textbf{BézierLaneNet (ResNet-18)} & \boldred{213} & \boldred{4.10} \\ %
        \textbf{BézierLaneNet (ResNet-34)} & 150 & 9.49 \\ %
        \bottomrule
    \end{tabular}}
    \caption{FPS (\textit{image/s}) and model size. All FPS results are tested with $360 \times 640$ random inputs on the same platform. Here only shows models with $>70\%$ CULane \cite{pan2018spatial} F1 score.} %
    \label{tab:profile}
    \vspace{-3mm}
\end{table}

To summarize, previous curve-based methods (PolyLaneNet \cite{tabelini2021polylanenet}, LSTR \cite{liu2021end}) have significantly worse performance. Fast methods trades either accuracy (UFLD \cite{qin2020ultra}) or model size (UFLD \cite{qin2020ultra}, LaneATT \cite{tabelini2021keep}) for speed. Accurate methods either discards the end-to-end pipeline (LaneATT \cite{tabelini2021keep}), or entirely fails the real-time requirement (SCNN \cite{pan2018spatial}, RESA \cite{zheng2021resa}). While our BézierLaneNet is fully end-to-end, fast (\textgreater 150 FPS), light-weight (\textless 10 million parameters) and maintains consistent high accuracy across datasets.

\subsection{Analysis}
\label{sec:analy}

Although we develop our method by tuning on the \textit{val} set, we re-run ablation studies with ResNet-34 backbone (including our full method) and report performance on the CULane \textit{test} set for clear comparison.

\begin{table}[ht]
    \centering
    \resizebox{6cm}{!}{\begin{tabular}{lC}
    \toprule
         \textbf{Curve representation} & \textbf{F1} \\
         \toprule
         Cubic Bézier curve baseline & 68.89 \\
         $3$rd Polynomial baseline & 1.49 \\
         \midrule
         BézierLaneNet & 75.41 \\
         $3$rd Polynomial from BézierLaneNet & 5.01 \\
         \bottomrule
    \end{tabular}}
    \caption{Curve representations. Baselines directly predict curve coefficients without feature flip fusion.}
    \label{tab:ablrep}
    \vspace{-4mm}
\end{table}

\noindent \textbf{Importance of Parametric Bézier Curve.} We first replace the Bézier curve prediction with a $3$rd order polynomial, adding auxiliary losses for start and end points.
As shown in \Cref{tab:ablrep}, polynomials catastrophically fail to converge in our fully convolutional network, even when trained with $150$ epochs (details in \Cref{subsec:bezierlanenet}).
Then we consider modifying the LSTR \cite{liu2021end} to predict cubic Bézier curves, the performance is similar to predicting polynomials. We conclude that heavy MLP may be necessary to learn polynomials \cite{tabelini2021polylanenet,liu2021end}, while predicting Bézier control points from position-aware CNN is the best choice. The transformer-based LSTR decoder destroys the fine spatial information, suppresses the advancement of curve function.

\begin{table}[ht]
    \centering
    \resizebox{6cm}{!}{\begin{tabular}{cccccC}
    \toprule
          \textbf{CP} & \textbf{SP} &\textbf{Flip} & \textbf{Deform} & \textbf{Seg} & \textbf{F1} \\
    \toprule
        \checkmark & & & & & 63.74 \\
        & \checkmark & & & & 68.89 \\
        & \checkmark &  & & \checkmark & 65.82 \\
        & \checkmark & \checkmark & & & 70.28 \\
        & \checkmark &  \checkmark & \checkmark & & 72.96 \\
        & \checkmark &  \checkmark & & \checkmark & 73.97 \\
        & \checkmark &  \checkmark & \checkmark & \checkmark & \mathbf{75.41}\\
         
    \bottomrule
         
    \end{tabular}}
    \caption{Ablations. \textbf{CP}: Control point loss \cite{liu2020abcnet}. \textbf{SP}: The proposed sampling loss. \textbf{Flip}: The feature flip fusion module. \textbf{Deform}: Employ the deformable convolution in feature flip fusion. \textbf{Seg}: Auxiliary segmentation loss.}
    \label{tab:abl}
    \vspace{-4mm}
\end{table}

\noindent \textbf{Feature Flip Fusion Design.} As shown in \Cref{tab:abl}, feature flip fusion brings $4.07\%$ improvement. We also find that the auxiliary segmentation loss can regularize and increase the performance further, by $2.45\%$. It is worth noting that auxiliary loss only works with feature fusion, it can lead to degenerated results when directly applied on the baseline ($-3.07\%$). A standard $3\times3$ convolution performs worse than deformable convolution, by $2.68\%$ and $1.44\%$, before and after adding the auxiliary segmentation loss, respectively. We attribute this to the effects of feature alignment.

\noindent \textbf{Bézier Curve Fitting Loss.} As shown in \Cref{tab:abl}, replacing the sampling loss by direct loss on control points lead to inferior performance ($-5.15\%$ in the baseline setup). Inspired by the success of IoU loss in object detection. We also implemented a IoU loss (formulas in \Cref{sec:iou}) for the convex hull of Bézier control points. However, the convex hull of close-to-straight lane lines are too small, the IoU loss is numerically unstable, thus failing to facilitate the sampling loss.

\begin{table}[ht]
    \centering
    \resizebox{7cm}{!}{\begin{tabular}{lcC}
    \toprule
         \textbf{Model} & \textbf{Aug} & \textbf{F1}  \\
    \toprule
         LSTR (ResNet-18, $2\times$) \cite{liu2021end} & \checkmark & 68.72 \\
         LSTR (ResNet-18, $2\times$) \cite{liu2021end} &  & 39.77 (-28.95) \\
         \midrule
         BézierLaneNet (ResNet-34) & \checkmark & 75.41\\ 
         BézierLaneNet (ResNet-34) &  & 55.11 (-20.30) \\
    \bottomrule
    \end{tabular}}
    \caption{Augmentation ablations. \textbf{Aug}: Strong data augmentation.}
    \label{tab:ablaug}
    \vspace{-3mm}
\end{table}

\noindent \textbf{Importance of Strong Data Augmentation.} Strong data augmentation is defined by a series of affine transforms and color distortions, the exact policy may slightly vary for different methods. For instance, we use random affine transform, random horizontal flip, and color jitter. LSTR \cite{liu2021end} also uses random lighting. Default augmentation includes only a small rotation ($3$ degrees). As shown in \Cref{tab:ablaug}, strong augmentation is essential to avoid over-fitting for curve-based methods.

For segmentation-based methods \cite{deeplabv1,pan2018spatial,zheng2021resa}, we fast validated strong augmentation on the smaller TuSimple \cite{tusimple} dataset. All shows a $1\sim2\%$ degradation. This suggests that they may be robust due to per-pixel prediction and heuristic post-processing. But they highly rely on learning the distribution of local features such as texture, which could become confusing by strong augmentation.

\subsection{Limitations and Discussions}
\label{sec:limit}

Curves are indeed a natural representation of lane lines. However, their elegance in modeling inevitably brings a drawback. It is difficult for the curvature coefficients to generalize when the data distribution is highly biased (almost all lane lines are straight lines in CULane). Our Bézier curve approach has already alleviated this problem to some extent and has achieved an acceptable performance ($62.45$) in CULane \textbf{Curve}. On datasets such as TuSimple and LLAMAS \cite{tusimple,llamas2019},
where the curvature distribution is fair enough for learning,
our method achieves even better performance. To handle broader corner cases, \eg, sharp turns, blockages and bad weather, datasets such as \cite{xu2020curvelane,Tan_2021_TIP_NightCity,sakaridis2018semantic} may be useful.

The feature flip fusion is specifically designed for a front-mounted camera, which is the typical use case of deep lane detectors. Nevertheless, there is still a strong inductive bias by assuming scene symmetry. In future work, it would be interesting to find a replacement for this module, to achieve better generalization and to remove the deformable convolution operation, which poses difficulty for effective integration into edge devices such as Jetson.

More discussions in \Cref{sec:dis}.

\section{Conclusions}
\label{sec:concl}

In this paper, we have proposed BézierLaneNet: a novel fully end-to-end lane detector based on parametric Bézier curves. The on-image Bézier curves are easy to optimize and naturally model the continuous property of lane lines, without heavy designs such as recurrent feature aggregation or heuristic anchors. Besides, a feature flip fusion module is proposed. It efficiently models the symmetry property of the driving scene, while also being robust to slight asymmetries by using deformable convolution. The proposed model has achieved favorable performance on three datasets, defeating all existing methods on the popular LLAMAS benchmark. It is also both fast (\textgreater $150$ FPS) and light-weight (\textless $10$ million parameters).

\noindent \textbf{Acknowledgements.} This work has been sponsored by National Key Research and Development Program of China (2019YFC1521104), National Natural Science Foundation of China (61972157, 72192821), Shanghai Municipal Science and Technology Major Project (2021SHZDZX0102), Shanghai Science and Technology Commission (21511101200), Art major project of National Social Science Fund (I8ZD22), and SenseTime Collaborative Research Grant. We thank Jiaping Qin for guidance on road design and geometry, Yuchen Gong and Pan Chen for helping with CAM visualizations, Zhijun Gong, Jiachen Xu and Jingyu Gong for insightful discussions about math, Fengqi Liu for providing GPUs, Lucas Tabelini for cooperation in evaluating \cite{tabelini2021polylanenet,tabelini2021keep}, and the CVPR reviewers for constructive comments.

{\small
\bibliographystyle{ieee_fullname}
\bibliography{egbib}
}

\clearpage

\appendix

\noindent \textbf{Appendix Overview.} The Appendix is organized as follows: \Cref{sec:fps} describes the FPS test protocol and environments; \Cref{sec:specs} introduces implementation details for each compared method (including ours in \Cref{subsec:bezierlanenet}); \Cref{sec:transforms} provides implementation details for Bézier curves, including sampling, ground truth generation and transforms; \Cref{sec:iou} formulates the IoU loss for Bézier curves and discusses why it failed; \Cref{sec:match} explores matching priors other than the \textit{centerness} prior; %
\Cref{sec:extra} shows extra ablation studies on datasets other than CULane \cite{pan2018spatial}, to verify the generalization of feature flip fusion; \Cref{sec:dis} discusses limitations and recognizes new progress in the field; \Cref{sec:qualitative} presents qualitative results from our method, visualized on three datasets.

\section{FPS Test Protocol}
\label{sec:fps}

Let one Frames-Per-Second (FPS) test trial be the average runtime of 100 consecutive model inference with its PyTorch \cite{paszke2019pytorch} implementation, without calculating gradients. The input is a $3$x$360$x$640$ random Tensor (some use all 1 \cite{tabelini2021keep}, which does not have impact on speed). Note that all methods do \textbf{not} use optimization from packages like TensorRT. We wait for all CUDA kernels to finish before counting the whole runtime. We use Python \textit{time.perf\_counter()} since it is more precise than \textit{time.time()}. For all methods, the FPS is reported as the best result from 3 trials.

Before each test trial, at least 10 forward pass is conducted as warm-up of the device. For each new method to be tested, we keep running warm-up trials of a recorded method until the recorded FPS is reached again, so we can guarantee a similar peak machine condition as before.

\vspace{1mm}

\noindent \textbf{Evaluation Environment.} The evaluation platform is a 2080 Ti GPU (standard frequency), on a Intel Xeon-E3 CPU server, with CUDA 10.2, CuDNN 7.6.5, PyTorch 1.6.0. FPS is a platform-sensitive metric, depending on GPU frequency, condition, bus bandwidth, software versions, \etc. Also using 2080 Ti, Tabelini \etal \cite{tabelini2021keep} can achieve a better peak performance for all methods. Thus we use the same platform for all FPS tests, to provide fair comparisons.

\vspace{1mm}

\noindent \textbf{Remark.} Note that \textbf{FPS (image/s)} is different from \textbf{throughput (image/s)}. Since FPS restricts batch size to $1$, which better simulates the real-time application scenario. While throughput considers a batch size more than $1$. LSTR \cite{liu2021end} reported a $420$ FPS for its fastest model, which is actually throughput with batch size $16$. Our re-tested FPS is $98$.

\section{Specifications for Compared Methods}
\label{sec:specs}

\subsection{Segmentation Baseline}
\label{subsec:segbase}

The segmentation baseline is based on DeeplabV1 \cite{deeplabv1}, originally proposed in SCNN \cite{pan2018spatial}. It is essentially the original DeeplabV1 without CRF, lanes are considered as different classes, and a separate lane existence branch (a series of convolution, pooling and MLP) is used to facilitate lane post-processing. We optimized its training and testing scheme based on recent advances \cite{zheng2021resa}. Re-implemented in our codebase, it attains higher performance than what recent papers usually report.

\noindent \textbf{Post-processing.} First, the existence of a lane is determined by the lane existence branch. Then, the predicted per-pixel probability map is interpolated to the input image size. After that, a $9 \times 9$ Gaussian blur is applied to smooth the predictions. Finally, for each existing lane class, the smoothed probability map is traversed by pre-defined Y coordinates (quantized), and corresponding X coordinates are recorded by the maximum probability position on the row (provided it passes a fixed threshold). Lanes with less than two qualified points are simply discarded.

\noindent \textbf{Data Augmentation.} We use a simple random rotation with small angles ($3$ degrees), then resize to input resolution.

\subsection{SCNN}
\label{subsec:scnn}

Our SCNN \cite{pan2018spatial} is re-implemented from the Torch7 version of the official code. Advised by the authors, we added an initialization trick for the spatial CNN layers, and learning rate warm-up, to prevent gradient explosion caused by recurrent feature aggregation. Thus, we can safely adjust the learning rate. Our improved SCNN achieves significantly better performance than the original one.

Some may find reports of $96.53$ accuracy of SCNN on TuSimple. However, that was a competition entry trained with external data. We report SCNN with ResNet backbones, trained with the same data as other re-implemented methods in our codebase.

\noindent \textbf{Post-processing.} Same as \Cref{subsec:segbase}.

\noindent \textbf{Data Augmentation.} Same as \Cref{subsec:segbase}.

\subsection{RESA}
\label{subsec:resa}

Our RESA \cite{zheng2021resa} is implemented based on its published paper. A main difference to the official code release is we do not cutout no-lane areas (in each dataset, there is a certain height range for lane annotation). Because that trick is dataset specific and not generalizable, we do not use that for all compared methods. Other differences are all validated to have better performance than the official code, at least on the CULane \textit{val} set.

\noindent \textbf{Post-processing.} Same as \Cref{subsec:segbase}.

\noindent \textbf{Data Augmentation.} Same as \Cref{subsec:segbase}. The original RESA paper \cite{zheng2021resa} also apply random horizontal flip, which was found ineffective in our re-implementation.

\subsection{UFLD}
\label{subsec:ufld}

Ultra Fast Lane Detection (UFLD) \cite{qin2020ultra} is reported from their paper and open-source code. Since TuSimple FP and FN information is not in the paper, and training from source code leads to very high FP rate (almost $20\%$), we did not report their performance on this dataset. We adjusted its profiling scripts to calculate number of parameters and FPS in our standard.

\noindent \textbf{Post-processing.} Since this method uses gridding cells (each cell is equivalent to several pixels in a segmentation probability map), each point's X coordinate is calculated as the expectation of locations (cells from the same row), i.e. a weighted average by probability. Differently from segmentation post-processing, it is possible to be efficiently implemented.

\noindent \textbf{Data Augmentation.} Augmentations include random rotation and some form of random translation.

\subsection{PolyLaneNet}
\label{subsec:polylanenet}

PolyLaneNet \cite{tabelini2021polylanenet} is reported from their paper and open-source code. We added a profiling script to calculate number of parameters and FPS in our standard.

\noindent \textbf{Post-processing.} This method requires no post-processing.

\noindent \textbf{Data Augmentation.} Augmentations include large random rotation ($10$ degrees), random horizontal flip and random crop. They are applied with a probability of $\frac{10}{11}$.

\subsection{LaneATT}
\label{subsec:laneatt}

LaneATT \cite{tabelini2021keep} is reported from their paper and open-source code. We adjusted its profiling scripts to calculate parameters and FPS in our standard.

\noindent \textbf{Post-processing.} Non-Maximal Suppression (NMS) is implemented by a customized CUDA kernel. An extra interpolation of lanes by B-Spline is removed both in testing and profiling, since it is slowly executed on CPU and provides little improvement ($\sim0.2\%$ on CULane).

\noindent \textbf{Data Augmentation.} LaneATT uses random affine transforms including scale, translation and rotation. While it also uses random horizontal flip.

\noindent \textbf{Followup.} We did not have time to validate the re-implementation of LaneATT in our codebase, prior the submission deadline. Therefore, the LaneATT performance is still reported from the official code. Our re-implementation indicates that all LaneATT results are reproducible except for the ResNet-34 backbone on CULane, which is slightly outside the standard deviation range, but still reasonable.

\subsection{LSTR}
\label{subsec:lstr}

LSTR \cite{liu2021end} is re-implemented in our codebase. All ResNet backbone methods start from ImageNet \cite{krizhevsky2012imagenet} pre-training. While LSTR \cite{liu2021end} use $256$ channels ResNet-18 for CULane ($2\times$), $128$ channels for other datasets ($1\times$), which makes it impossible to use off-the-shelf pre-trained ResNets. Although whether ImageNet pre-training helps lane detection is still an open question. Our reported performance of LSTR on CULane, is the first documented report of LSTR on this dataset. With tuning of hyper-parameters (learning rate, epochs, prediction threshold), bug fix (the original classification branch has $3$ output channels, which should be $2$), we achieve $4\%$ better performance on CULane than the authors' trial. Specifically, we use learning rate $2.5\times10^{-4}$ with batch size $20$. $150$ and $2000$ epochs, $0.95$ and $0.5$ prediction thresholds, for CULane and TuSimple. The lower threshold in TuSimple is due to the official test metric, which significantly favors a high recall. However, for real-world applications, a high recall leads to high False Positive rate, which is undesired.

We divide the curve loss weighting by $10$ with our LSTR-Beizer ablation, since there were $100$ sample points with both X and Y coordinates to fit, that is a loss scale about $10$ times the original loss (LSTR loss takes summation of point $L1$ distances instead of average). This modulation achieves a similar loss landscape to original LSTR.

\noindent \textbf{Post-processing.} This method requires no post-processing.

\noindent \textbf{Data Augmentation.} Data augmentation includes PolyLaneNet's (\Cref{subsec:polylanenet}), then appends random color distortions (brightness, contrast, saturation, hue) and random lighting by a light source calculated from the COCO dataset \cite{lin2014microsoft}. That is by far the most complex data augmentation pipeline in this research field, we have validated that all components of this pipeline helps LSTR training.

\noindent \textbf{Remark.} The polynomial coefficients of LSTR are unbounded, which leads to numerical instability (while the bipartite matching requires precision), and high failure rate of training. The failure rate of fp$32$ training on CULane is $\sim 30\%$. This is circumvented in BézierLaneNet, since our $L1$ loss can be bounded to $[0,1]$ without influence on learning (control points easily converges to on-image).

\subsection{BézierLaneNet}
\label{subsec:bezierlanenet}

BézierLaneNet is implemented in the same code framework where we re-implemented other methods. Same as LSTR, the default prediction threshold is set to $0.95$, while $0.5$ is used for TuSimple \cite{tusimple}.

\noindent \textbf{Post-processing.} This method requires no post-processing.

\noindent \textbf{Data Augmentation.} We use augmentations similar to LSTR (\Cref{subsec:lstr}). Concretely, we remove the random lighting from LSTR (to strictly avoid using knowledge from external data), and replace the PolyLaneNet $\frac{10}{11}$ chance augmentations with random affine transforms and random horizontal flip, like LaneATT (\Cref{subsec:laneatt}). The random affine parameters are: rotation ($10$ degrees), translation (maximum $50$ pixels on X, $20$ on Y), scale (maximum $20\%$).

\noindent \textbf{Polynomial Ablations.} For the polynomial ablations (Table 7), we modified the network to predict 6 coefficients for $3$rd order Polynomial ($4$ curve coefficients and start/end Y coordinates). Extra $L1$ losses are added for the start/end Y coordinates similar to LSTR \cite{liu2021end}. With extensive tryouts (adjusting learning rate, loss weightings, number of epochs), even at the full BézierLaneNet setup, with $150$ epochs on CULane, the models still can not converge to a good enough solution. In other word, not precise enough to pass the CULane metric. The sampling loss on polynomial curves can only get to $0.02$, which means $0.02 \times 1640 \mathrm{pixels} = 32.8 \mathrm{pixels}$ average X coordinate error on training set. 
CULane requires a $0.5$ IoU between curves, which are enlarged to $30$ pixels wide, thus at least around $10$ pixels average error is needed to get meaningful results. By loosen up the IoU requirement to $0.3$, we can get F1 score $15.82$ for ``$3$rd Polynomial from BézierLaneNet''. Although the reviewing committee suggested adding simple regularization for this ablation to converge, regretfully we failed to do this.

\section{Bézier Curve Implementation Details}
\label{sec:transforms}

\noindent \textbf{Fast Sampling.} The sampling of Bézier curves may seem tiresome due to the complex Bernstein basis polynomials. To fast sample a Bézier curve by a series of fixed $t$ values, simply pre-compute the results from Bernstein basis polynomials, thus only one simple matrix multiplication is left.

\noindent \textbf{Remarks on GT Generation.} The ground truth of Bézier curves are generated with least squares fitting, a common technique for polynomials. We use it for its simplicity and the fact that it already shows near-perfect lane line fitting ability ($99.996$ and $99.72$  F1 score on CULane \textit{test} and LLAMAS \textit{val}, respectively). However, it is not an ideal algorithm for parametric curves. There is a whole research field for fitting Bézier curves better than least squares \cite{pastva1998bezier}.

\noindent \textbf{Bézier Curve Transform.} Another implementation difficulty on Bézier curves is how to apply affine transform (for transforming ground truth curves in data augmentation). Mathematically, affine transform on the control points is equivalent to affine transform on the entire curve. However, translation or rotation can move control points out of the image. In this case, a cutting of Bézier curves is required. The classical De Casteljau's algorithm is used for cutting an on-image Bézier curve segment. Assume a continuous on-image segment, valid sample points with minimum boundary $t = t_{0}$, maximum boundary $t = t_{1}$. The formula to cut a cubic Bézier curve defined by control points $\mathcal{P}_{0}, \mathcal{P}_{1}, \mathcal{P}_{2}, \mathcal{P}_{3}$ to its on-image segment $\mathcal{P}_{0}', \mathcal{P}_{1}', \mathcal{P}_{2}', \mathcal{P}_{3}'$, is derived as:

\vspace{-5mm}

\begin{equation}
\begin{split}
    \mathcal{P}_{0}' &=  u_{0}u_{0}u_{0}\mathcal{P}_{0} + (t_{0}u_{0}u_{0} + u_{0}t_{0}u_{0} + u_{0}u_{0}t_{0})\mathcal{P}_{1} \\
    & + (t_{0}t_{0}u_{0} + u_{0}t_{0}t_{0} + t_{0}u_{0}t_{0})\mathcal{P}_{2} + t_{0}t_{0}t_{0}\mathcal{P}_{3}, \\
    \mathcal{P}_{1}' &= u_{0}u_{0}u_{1}\mathcal{P}_{0} + (t_{0}u_{0}u_{1} + u_{0}t_{0}u_{1} + u_{0}u_{0}t_{1})\mathcal{P}_{1} \\
    &+ (t_{0}t_{0}u_{1} + u_{0}t_{0}t_{1} + t_{0}u_{0}t_{1}) \mathcal{P}_{2} + t_{0}t_{0}t_{1} \mathcal{P}_{3}, \\
    \mathcal{P}_{2}' &= u_{0}u_{1}u_{1} \mathcal{P}_{0} + (t_{0}u_{1}u_{1} + u_{0}t_{1}u_{1} + u_{0}u_{1}t_{1}) \mathcal{P}_{1} \\
    &+ (t_{0}t_{1}u_{1} + u_{0}t_{1}t_{1} + t_{0}u_{1}t_{1}) \mathcal{P}_{2} + t_{0}t_{1}t_{1} \mathcal{P}_{3}, \\
    \mathcal{P}_{3}' &= u_{1}u_{1}u_{1} \mathcal{P}_{0} + (t_{1}u_{1}u_{1} + u_{1}t_{1}u_{1} + u_{1}u_{1}t_{1}) \mathcal{P}_{1} \\
    &+ (t_{1}t_{1}u_{1} + u_{1}t_{1}t_{1} + t_{1}u_{1}t_{1}) \mathcal{P}_{2} + t_{1}t_{1}t_{1} \mathcal{P}_{3}, \\
\end{split}
\label{eq:beziercut}
\end{equation}
where $u_{0} = 1 - t_{0}$, $u_{1} = 1 - t_{1}$. This formula can be efficiently implemented by matrix multiplication. The possibility of noncontinuous cubic Bézier segment on lane detection datasets is extremely low and thus ignored for simplicity. If it does happen, \Cref{eq:beziercut} will not change the curve, while our network can also predict out-of-image control points, which still fit the on-image lane segments.

\section{IoU Loss for Bézier Curves}
\label{sec:iou}

Here we briefly introduce how we formulated the IoU loss between Bézier curves. Before diving into the algorithm, there are two preliminaries.

\begin{itemize}
\vspace{-2mm}
    \item Polar sort: By anchoring on an arbitrary point inside the N-sided polygon with vertices ${c_{i}(x_{i}, y_{i})}_{i=1}^{N}$ (normally the mean coordinate between vertices $c'=(\frac{1}{N}\sum_{i=1}^{N}x_{i}, \frac{1}{N}\sum_{i=1}^{N}y_{i})$), vertices are sorted by its \textit{atan2} angles. This will return a clockwise or counterclockwise polygon.
\vspace{-2mm}
    \item Convex polygon area: A sorted convex polygon can be efficiently cut into consecutive triangles by simple indexing operations. The convex polygon area is the sum of these triangles. The area $S$ of triangle $((x_1, y_1), (x_2, y_2), (x_3, y_3))$ is: $S = \frac{1}{2} | x_1(y_2 - y_3) + x_2(y_3 - y_1) + x_3(y_1 - y_2) |$.
\vspace{-2mm}
\end{itemize}

Assume we have two convex hulls from Bézier curves (there are a lot of convex hull algorithms).
Now the IoU between Bézier curves are converted to IoU between convex polygons. Based on the simple fact that the intersection of convex polygons is still a convex polygon, after polar sorting all the convex hulls and determining the intersected polygon, we can easily formulate IoU calculations as a series of convex polygon area calculations. The difficulty lies in how to efficiently determine the intersection between convex polygon pairs. 

Consider two intersected convex polygons, their intersection includes two types of vertices:

\begin{itemize}
    \item Intersections: intersection points between edges.
    \item Insiders: vertices inside/on both polygons.
\end{itemize}

For Intersections, we first represent every polygon edge as the general line equation: $ax + by = c$.
Then, for line $a_1x + b_1y = c_1$ and line $a_2x + b_2y = c_2$,
the intersection $(x', y')$ is calculated by:

\vspace{-3mm}

\begin{equation}
\begin{split}
    x' &= (b_2c_1 - b_1c_2) / det \\
    y' &= (a_1c_2 - a_2c_1) / det, \\
\end{split}
\label{eq:lineintersec}
\end{equation}
where $det = a_1b_2 - a_2b_1$. All $(x', y')$ that is on the respective line segments are Intersections.

For Insiders, there is a certain definition:

\begin{definition}
For a convex polygon, point $P(x, y)$ on the same side of each edge is inside the polygon.
\end{definition}

A sorted convex polygon is a series of edges (line segments defined by $P_0(x_0, y_0), P_1(x_1, y_1)$), the equation to decide which side a point is to a line segment is as follows:

\vspace{-3mm}

\begin{equation}
    sign = (y - y_0) (x_1 - x_0) - (x - x_0) (y_1 - y_0).
\end{equation}

$sign > 0$ means $P$ is on the right side, $sign < 0$ is the left side, and $sign = 0$ means $P$ is on the line segment. Note that equality is not a stable operation for float computations. But there are simple ways to circumvent that in coding, which we will not elaborate here.

There are other ways to determine Intersections and Insiders, but the above formulas can be efficiently implemented with matrix operations and indexing, making it possible to quickly train networks with batched inputs.

Finally, after being able to compute convex polygon intersections and areas, the Generalized IoU loss (GIoU) is simply (as in \cite{rezatofighi2019generalized}):

\begin{algorithm}[h]\label{algo:GIOU}	%
	\small{
	\SetKwInOut{Input}{input}\SetKwInOut{Output}{output}
		\Input{Two arbitrary convex shapes: $A,B\subseteq\mathbb{S}\in\mathbb{R}^n$}
		\Output{$GIoU$}}
		1. For $A$ and $B$, find the smallest enclosing convex object $C$, where $C\subseteq\mathbb{S}\in\mathbb{R}^n$ \\
		2. $\displaystyle IoU = \frac{|A\cap B|}{|A\cup B|}$\\
		3. $\displaystyle GIoU = IoU - \frac{|C\backslash(A\cup B)|}{|C|}$
\end{algorithm}

Union is computed as $A\cup B = A + B - A\cap B$. The enclosing convex object $C$ can be computed as the convex hull of two convex polygons, or upper-bounded by a enclosing rectangle. We implement the IoU computation purely in PyTorch \cite{paszke2019pytorch}, the runtime for our implementation is only about $5\times$ the runtime of rectangle IoU loss computation.

However, lane lines are mostly straight based on road design regulations \cite{american,chinese}. This leads to extremely small convex hull area for Bézier curves, thus introduces numerical instabilities in optimization. Although succeeded in a toy polygon fitting experiment, we currently failed to observe the loss's convergence to help learning on lane datasets.

\section{GT and Prediction Matching Prior}
\label{sec:match}

\begin{figure}[h]
\centering
    \begin{tabular}{c}
    \includegraphics[width = 0.99\linewidth, height=0.4cm]{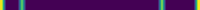}\\
    \includegraphics[width = 0.25\linewidth, height=0.4cm]{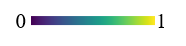}\\
    \end{tabular}
    \caption{Logits activation statistics ($1 \times \frac{W}{16}$) on CULane \cite{pan2018spatial}.}
\label{fig:localmaximum}
\end{figure}

Instead of the \textit{centerness} prior, we explore a local maximum prior, \ie, restricts matched prediction to have a local maximum classification logit. This prior can facilitate the model to understand the spatially sparse structure of lane lines. As shown in \Cref{fig:localmaximum}, the learned feature activation for classification logits exhibits a similar structure as an actual driving scene.

\def\imh{2.8cm}
\begin{figure*}[ht]
\renewcommand{\tabcolsep}{1pt}
\centering
    \begin{tabular}{ccc}
    \includegraphics[width = 0.45\linewidth, height=\imh]{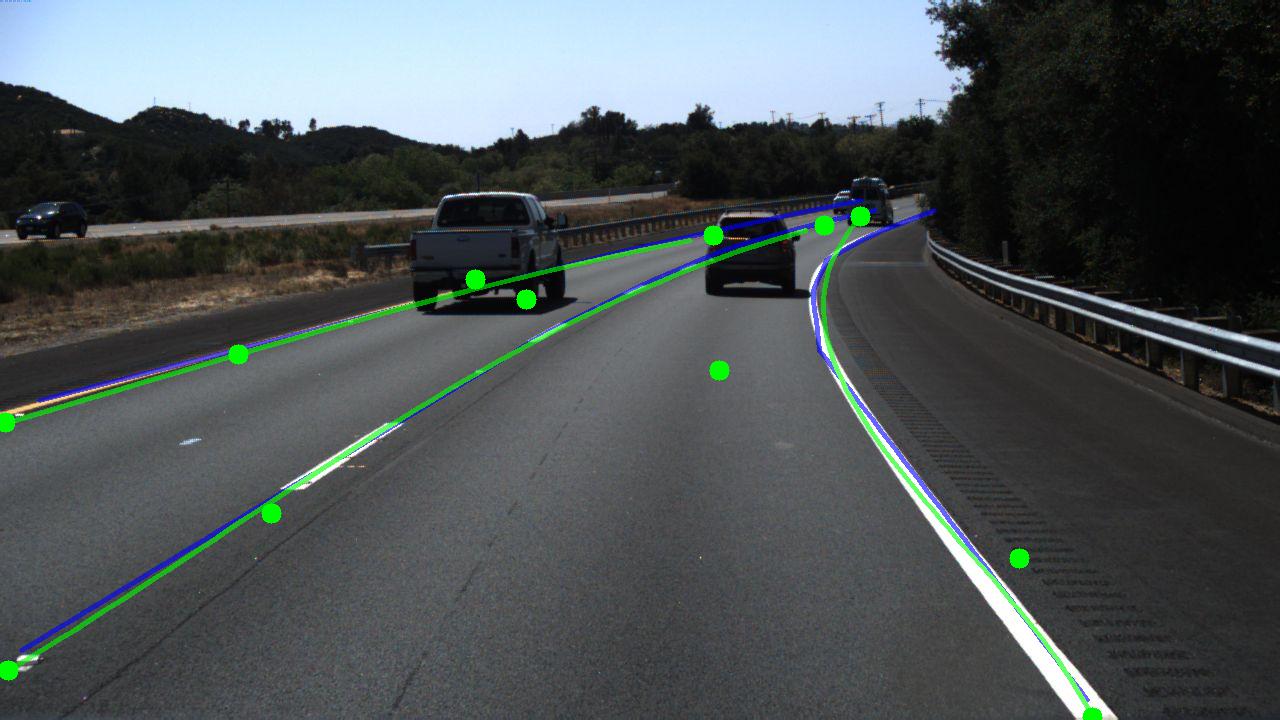}&&
    \includegraphics[width = 0.45\linewidth, height=\imh]{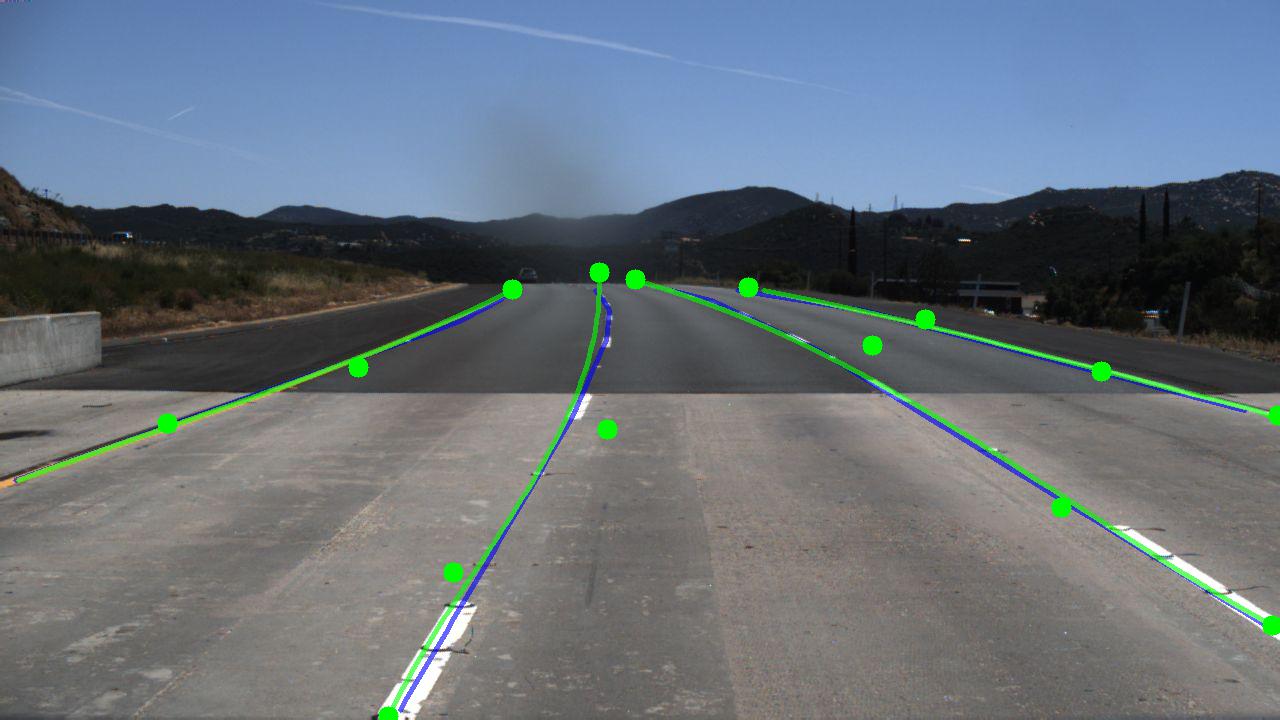}\\
    \includegraphics[width = 0.45\linewidth, height=\imh]{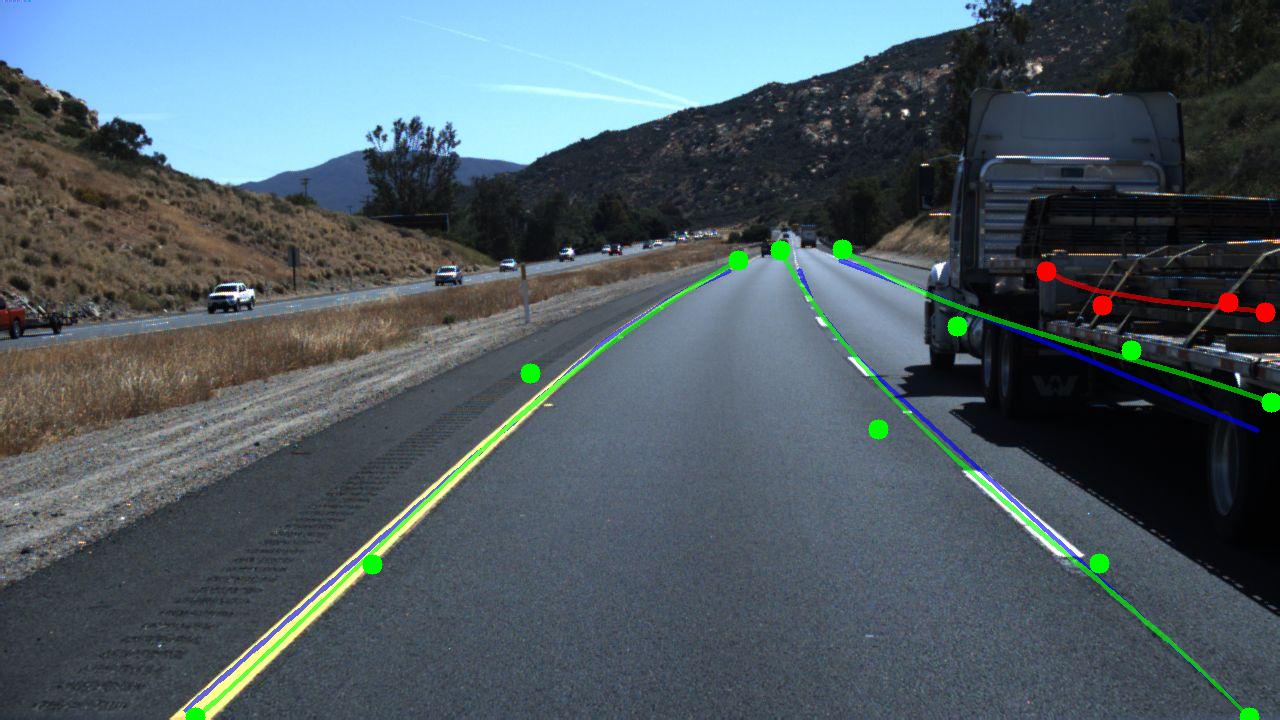}&&
    \includegraphics[width = 0.45\linewidth, height=\imh]{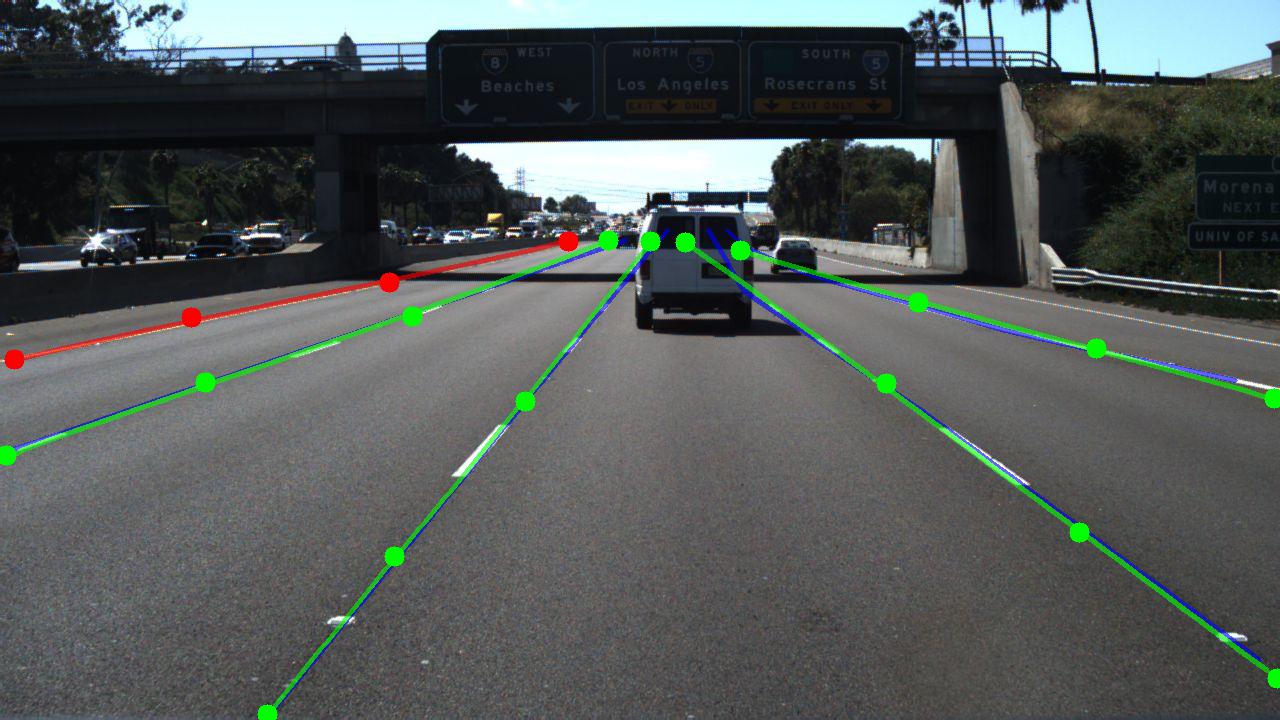}\\
    \multicolumn{3}{c}{(a) TuSimple \cite{tusimple}.}\\
    \includegraphics[width = 0.45\linewidth, height=\imh]{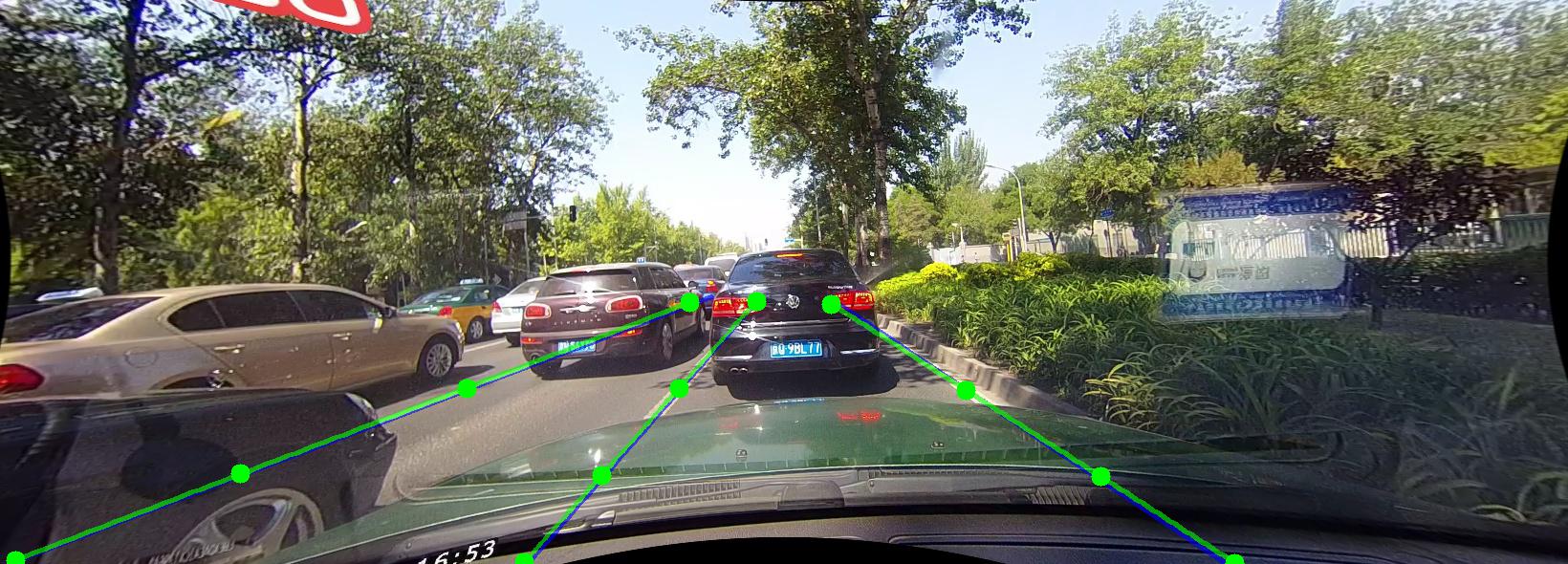}&&
    \includegraphics[width = 0.45\linewidth, height=\imh]{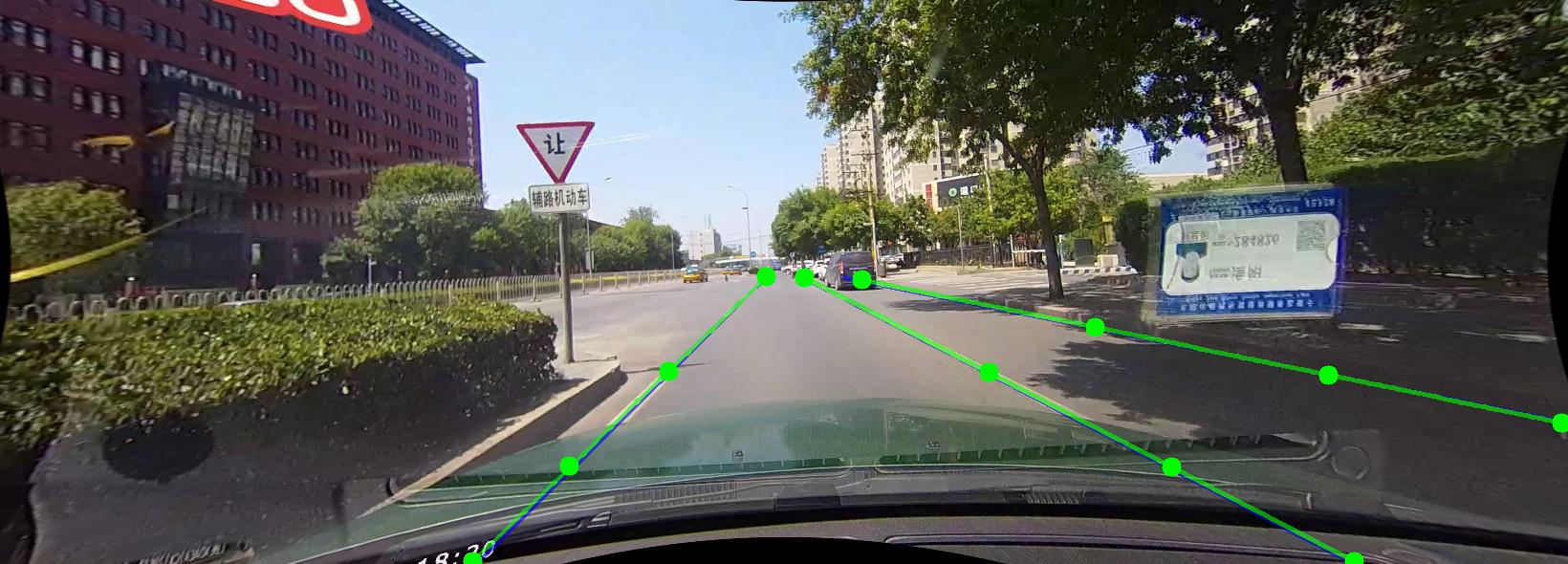}\\
    \includegraphics[width = 0.45\linewidth, height=\imh]{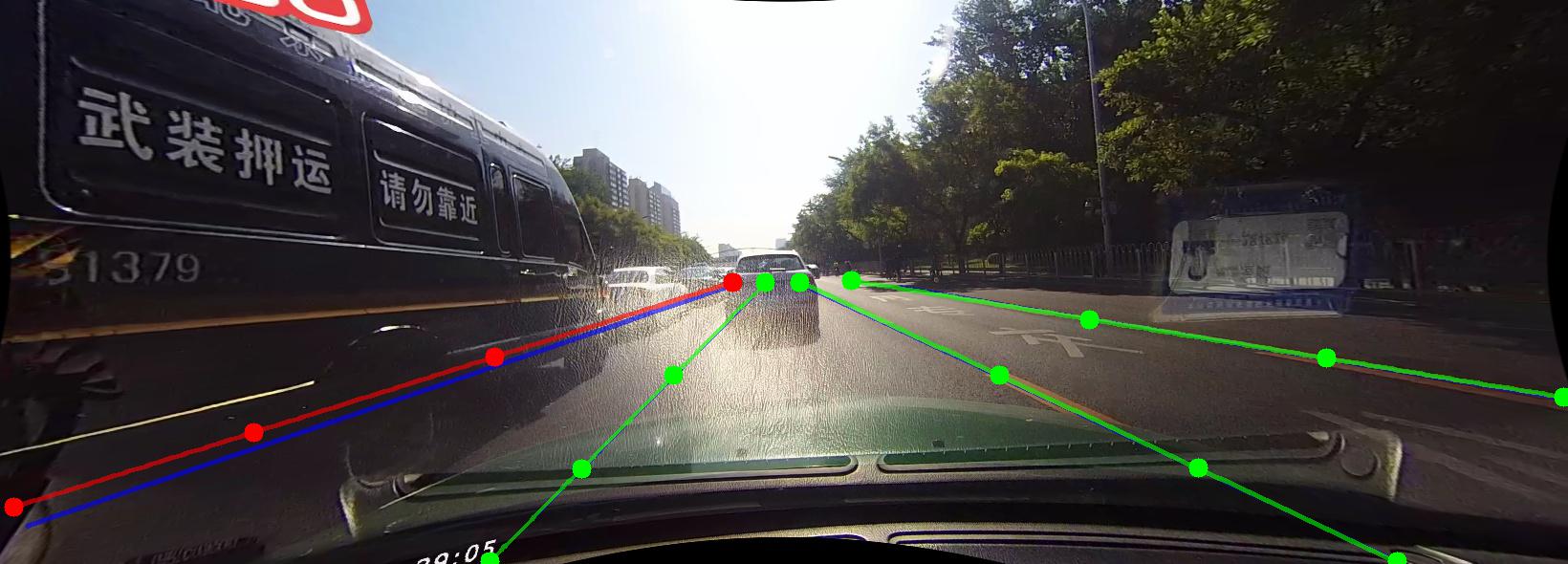}&&
    \includegraphics[width = 0.45\linewidth, height=\imh]{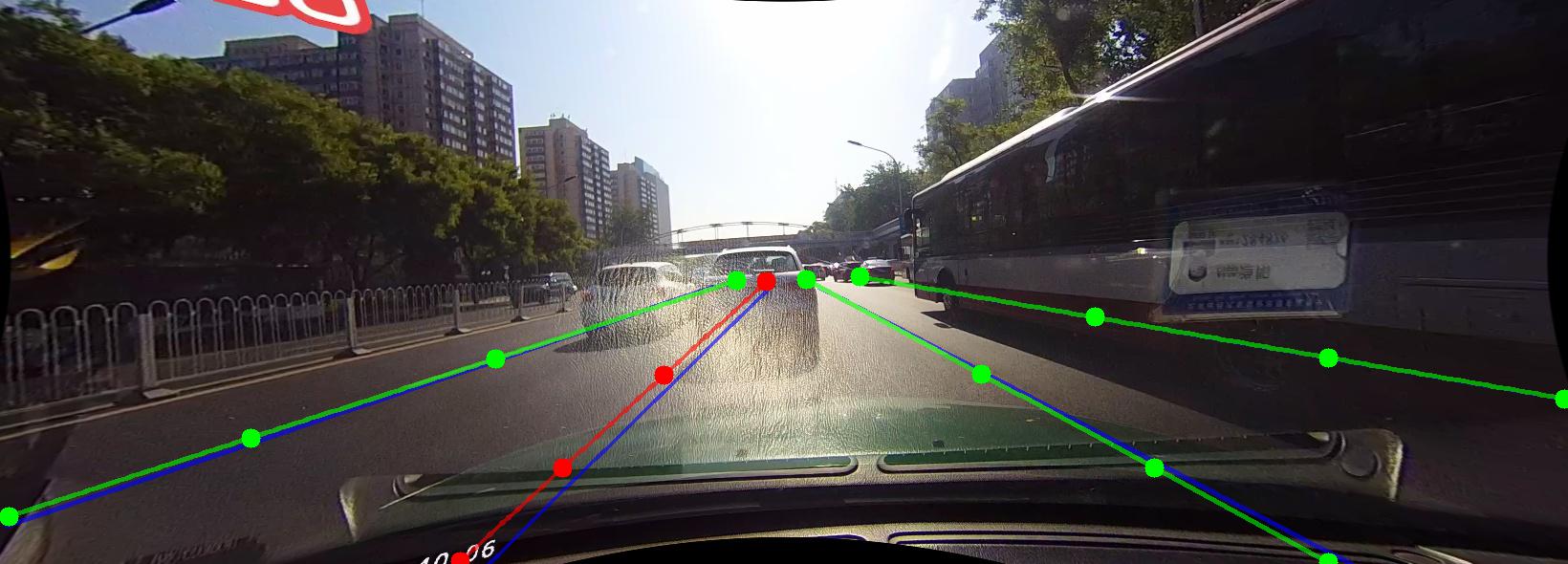}\\
    \multicolumn{3}{c}{(b) CULane \cite{pan2018spatial}.}\\
    \includegraphics[width = 0.45\linewidth, height=\imh]{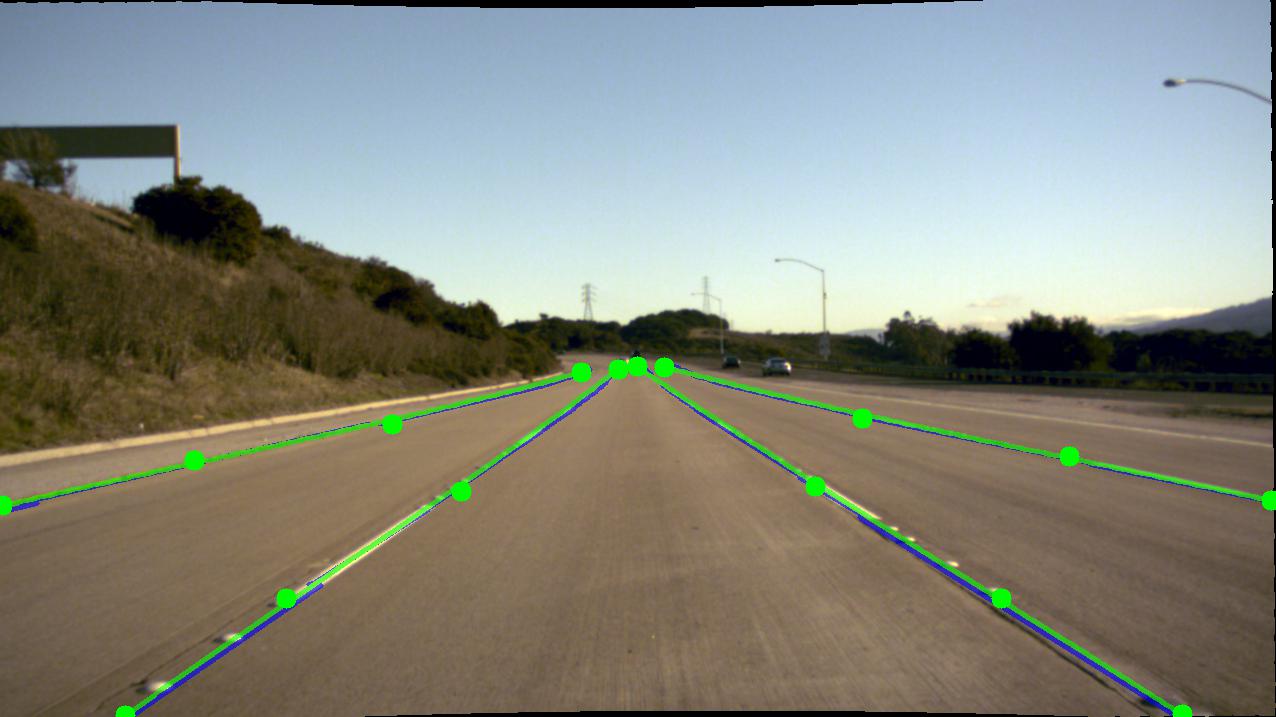}&&
    \includegraphics[width = 0.45\linewidth, height=\imh]{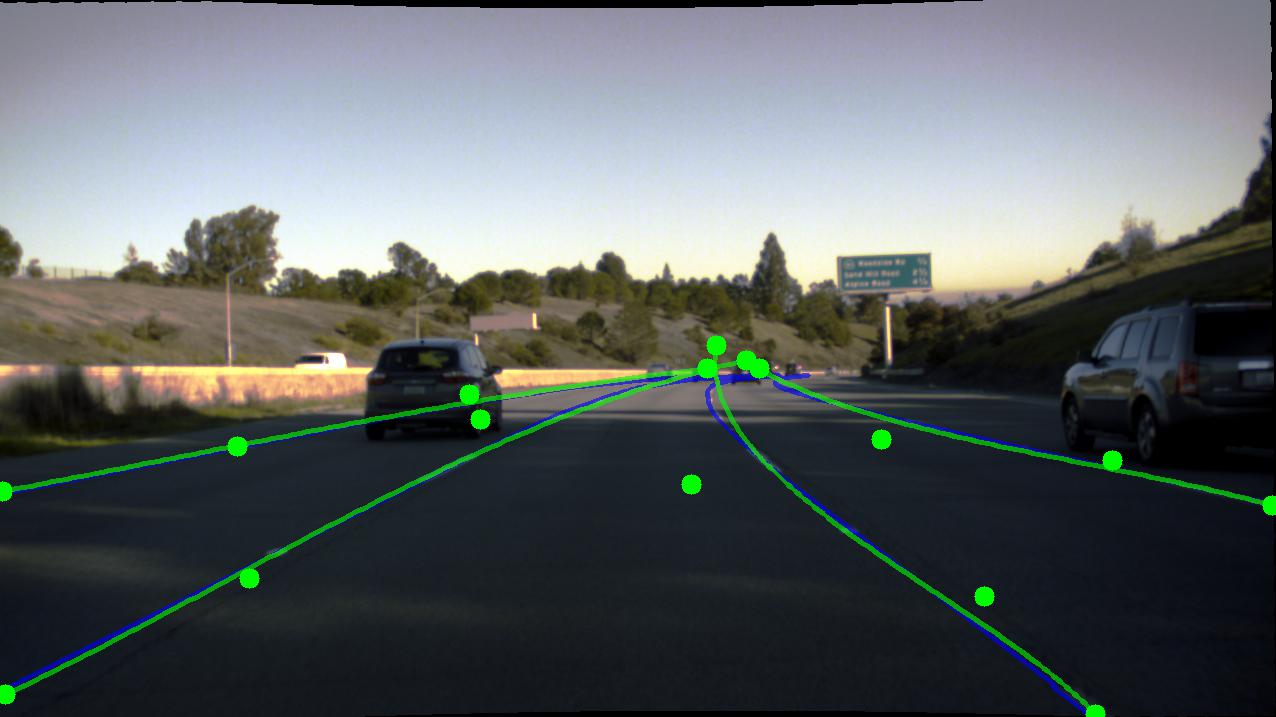}\\
    \includegraphics[width = 0.45\linewidth, height=\imh]{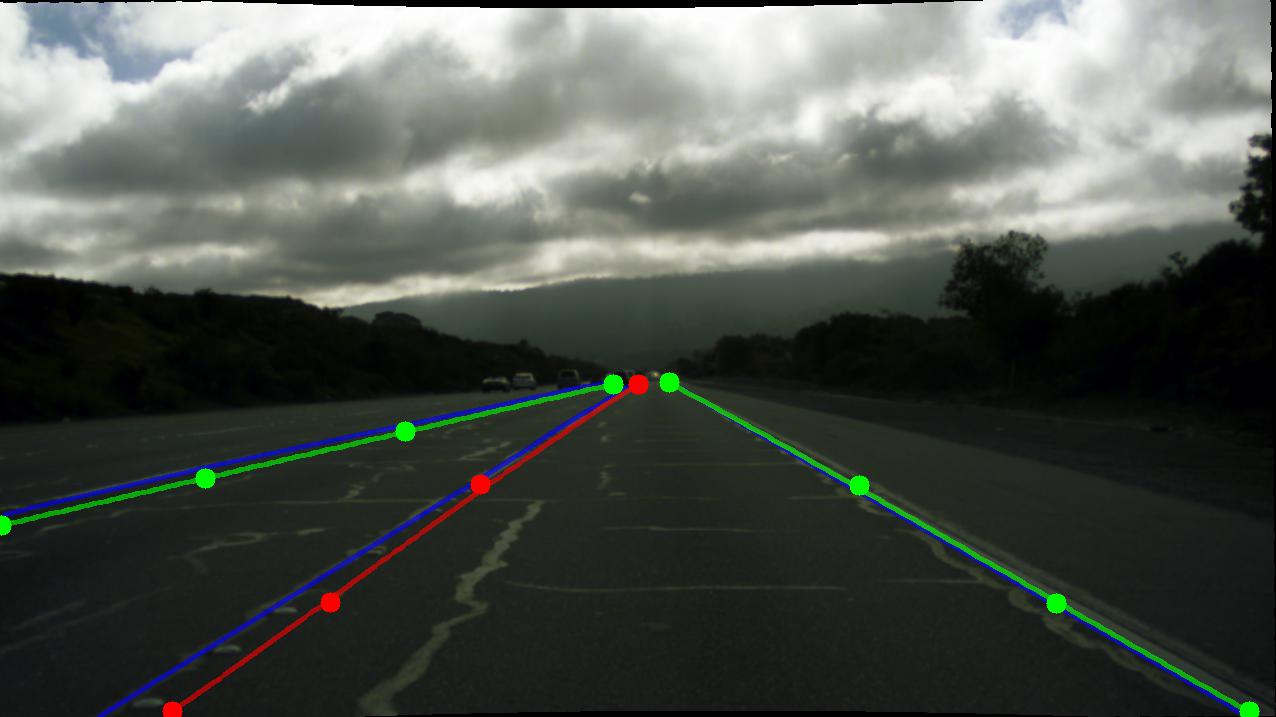}&&
    \includegraphics[width = 0.45\linewidth, height=\imh]{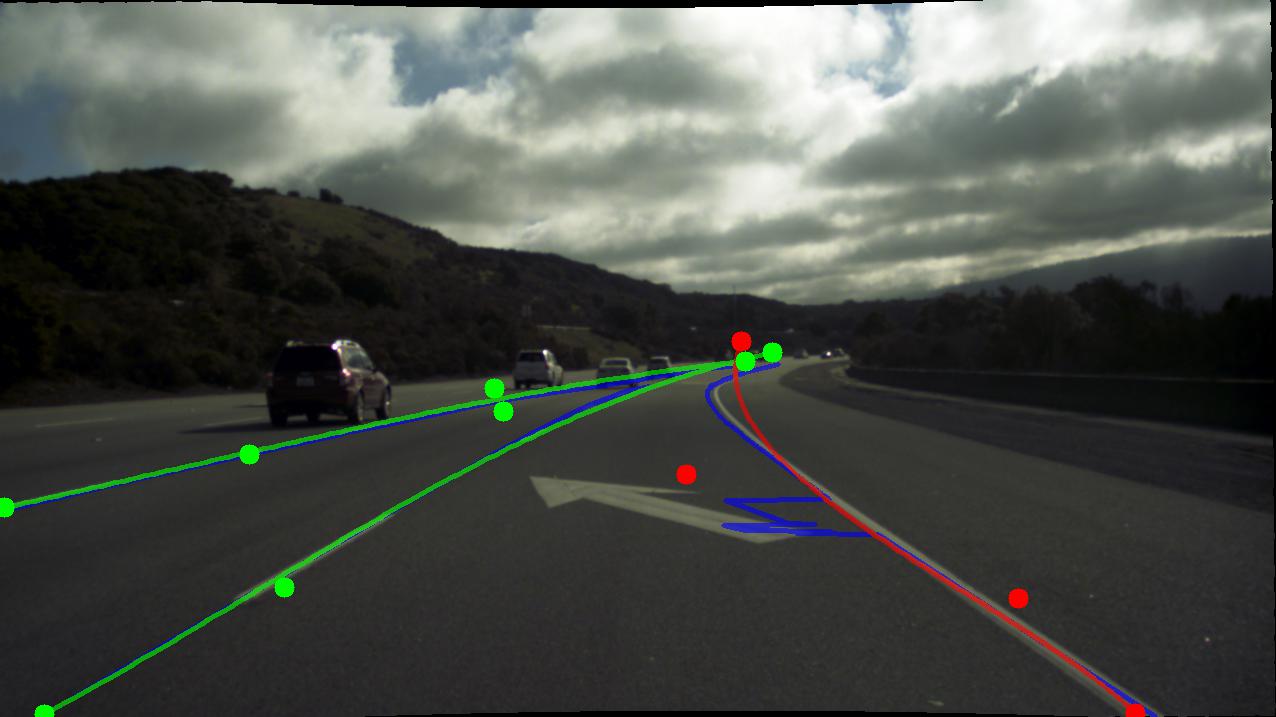}\\
    \multicolumn{3}{c}{(c) LLAMAS \cite{llamas2019}.}\\
    \end{tabular}
    \vspace{-1mm}
\caption{Qualitative results from BézierLaneNet (ResNet-34) on \textit{val} sets. False Positives (FP) are marked by red, True Positives (TP) are marked by green, ground truth are drawn in blue. Blue lines that are barely visible are precisely covered by green lines. Bézier curve control points are marked with solid circles. Images are slightly resized for alignment. Best viewed in color, in $2\times$ scale.}
\vspace{-5mm}

\label{fig:visall}
\end{figure*}

\section{Extra Results}
\label{sec:extra}

\begin{table}[h]
    \centering
    \resizebox{7cm}{!}{\begin{tabular}{lCC}
    \toprule
         & \textbf{TuSimple \cite{tusimple}} & \textbf{LLAMAS \cite{llamas2019}} \\
        \toprule
        Bézier Baseline & 93.36 & 95.27 \\
        + Feature Flip Fusion  & 95.26~(+ 1.90)  & 96.00~(+ 0.73) \\
        \bottomrule
    \end{tabular}}
    \caption{Ablation study on TuSimple (\textit{test} set Accuracy) and LLAMAS (\textit{val} set F1), before and after adding the Feature Flip Fusion module. Reported 3-times average with the ResNet-34 backbone, since ablations often are not stable enough on these datasets to exhibit a clear difference between methods.}
    \label{tab:moreabl}
    \vspace{-6mm}
\end{table}

\section{Discussions}
\label{sec:dis}

There exists a primitive application of lane detectors from lateral views to estimate the distance to the border of the drivable area \cite{gurghian2016deeplanes}, which contradicts the use of feature flip fusion. In this case, possibly a lower order Bézier curve baseline (with row-wise instead of column-wise pooling) would suffice. This is out of the focus of this paper.

\noindent \textbf{Recent Progress.} Recently, others have explored alternative lane representation or formulation methods that do not fully fit in the three categories (segmentation, point detection, curve). Instead of the popular top-down regime, \cite{qu2021focus} propose a bottom-up approach that focus on local details. \cite{liu2021condlanenet} achieve state-of-the-art performance, but the complex conditional decoding of lane lines results in unstable runtime depending on the input image, which is not desirable for a real-time system.

\section{Qualitative Results}
\label{sec:qualitative}

Qualitative results are shown in \Cref{fig:visall}, from our ResNet-34 backbone models.
For each dataset, 4 results are shown in two rows: first row shows qualitative successful predictions; second row shows typical failure cases.

\noindent \textbf{TuSimple.} As shown in \Cref{fig:visall}(a), our model fits highway curves well, only slight errors are seen on the far side where image details are destroyed by projection. Our typical failure case is a high FP rate, mostly attributed to the use of low threshold (\Cref{subsec:bezierlanenet}). However, in the bottom-right wide road scene, our FP prediction is actually a meaningful lane line that is ignored in center line annotations.

\noindent \textbf{CULane.} As shown in \Cref{fig:visall}(b), most lanes in this dataset are straight. Our model can make accurate predictions under heavy congestion (top-left) and shadows (top-right, shadow cast by trees).
A typical failure case is inaccurate prediction under occlusion (second row), in these cases one often cannot visually tell which one is better (ground truth or our FP prediction).

\noindent \textbf{LLAMAS.} As shown in \Cref{fig:visall}(c), our method performs accurate for clear straight-lines (top-left), and also good for large curvatures in a challenging scene almost entirely covered by shadow. In bottom-left image, our model fails in a low-illumination, tainted road. While in the other low-illumination scene (bottom-right), the unsupervised annotation from LIDAR and HD-map is misled by the white arrow (see the zigzag shape of the right-most blue line).

\end{document}